\documentclass[letterpaper]{article} 
\usepackage{aaai2026}  
\usepackage{times}  
\usepackage{helvet}  
\usepackage{courier}  
\usepackage[hyphens]{url}  
\usepackage{graphicx} 
\usepackage{natbib}  
\usepackage{caption} 
\usepackage{amsmath}
\usepackage{amssymb}
\usepackage{booktabs}
\usepackage{multirow} 
\usepackage{subcaption}
\usepackage{array}
\usepackage{amsthm}
\newtheorem{theorem}{Theorem}
\newtheorem{definition}{Definition}
\usepackage{makecell}
\usepackage{adjustbox}
\usepackage{algorithm}
\usepackage{algpseudocode}

\usepackage{newfloat}
\usepackage{listings}
\DeclareCaptionStyle{ruled}{labelfont=normalfont,labelsep=colon,strut=off} 
\floatstyle{ruled}
\newfloat{listing}{tb}{lst}{}
\floatname{listing}{Listing}
\title{FreDN: Spectral Disentanglement for Time Series Forecasting via Learnable Frequency Decomposition}
\author{
    Zhongde An\textsuperscript{\rm 1},
    Jinhong You\textsuperscript{\rm 1},
    Jiyanglin Li\textsuperscript{\rm 2},\\
    Yiming Tang\textsuperscript{\rm 3},
    Wen Li\textsuperscript{\rm 4}\thanks{Corresponding authors},
    Heming Du\textsuperscript{\rm 5}\footnotemark[1],
    Shouguo Du\textsuperscript{\rm 6}
}
\affiliations{
    \textsuperscript{\rm 1}Shanghai University of Finance and Economics, Shanghai, China\\
    \textsuperscript{\rm 2}Guizhou University of Finance and Economics, Guiyang, China\\
    \textsuperscript{\rm 3}Shanghai Lixin University of Accounting and Finance, Shanghai, China\\
    \textsuperscript{\rm 4}Shanghai University of International Business and Economics, Shanghai, China\\
    \textsuperscript{\rm 5}Australian National University, Canberra, Australia\\
    \textsuperscript{\rm 6}Shanghai Municipal Big Data Center, Shanghai, China\\
    \{2023213372@stu.sufe.edu.cn, 
    johnyou07@163.com, 
    jylli@mail.gufe.edu.cn, 
    jstangyiming@163.com, 
    wen.li.sh@outlook.com, 
    heming.du@anu.edu.au, 
    shouguo.du.sh@outlook.com\}
}

\begin{document}

\maketitle

\begin{abstract}
Time series forecasting is essential in a wide range of real world applications. Recently, frequency-domain methods have attracted increasing interest for their ability to capture global dependencies. However, when applied to non-stationary time series, these methods encounter the $\textit{spectral entanglement}$ and the computational burden of complex-valued learning. The $\textit{spectral entanglement}$ refers to the overlap of trends, periodicities, and noise across the spectrum due to $\textit{spectral leakage}$ and the presence of non-stationarity. However, existing decompositions are not suited to resolving spectral entanglement. To address this, we propose the Frequency Decomposition Network (FreDN), which introduces a learnable Frequency Disentangler module to separate trend and periodic components directly in the frequency domain. Furthermore, we propose a theoretically supported ReIm Block to reduce the complexity of complex-valued operations while maintaining performance. We also re-examine the frequency-domain loss function and provide new theoretical insights into its effectiveness. Extensive experiments on seven long-term forecasting benchmarks demonstrate that FreDN outperforms state-of-the-art methods by up to 10\%. Furthermore, compared with standard complex-valued architectures, our real-imaginary shared-parameter design reduces the parameter count and computational cost by at least 50\%. We have made our code publicly available at https://github.com/An-z-d/FreDN
\end{abstract}

\section{Introduction}
Time series forecasting is essential in energy, finance, and traffic management. Time-domain forecasting models continue to make notable progress in recent years. MLP-based architectures like TimeMixer \cite{wang2024timemixer} and SOFTS \cite{han2024softs} leverage multiscale mixing or centralized channel fusion to improve efficiency and scalability. At the same time, frequency-domain methods have gained traction for their ability to capture global dependencies and suppress noise. Models like FreTS \cite{yi2023frequency} suggest that the frequency spectrum offers MLPs a more complete view of the signal, facilitating the learning of global patterns. Recent works such as FITS \cite{xu2024fits} and FreDF \cite{wang2025fredf} further demonstrate that frequency-based representations enhance forecasting accuracy.

\begin{figure*}
    \centering
    \includegraphics[width=0.95\textwidth]{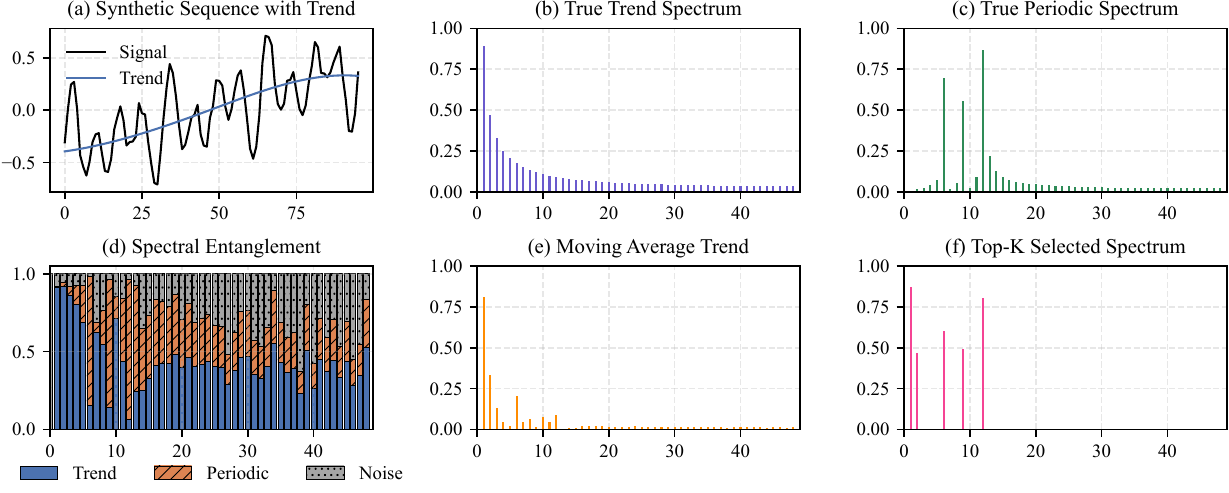}
    \caption{
        Spectral component entanglement.
        Top row: (a) the synthetic sequence with trend generated by B-spline; (b) the frequency spectrum of the real trend; (c) the frequency spectrum of the real periodicity. 
        Bottom row: (d) the frequency-wise proportion of trend, periodic, and noise in the original synthetic sequence; (e) the spectrum of the trend extracted by the moving average in the time domain; (f) the spectrum of the TopK-selected frequency components.
    }
    \label{fig:trend}
\end{figure*}
Frequency-domain techniques have become increasingly prevalent in time series forecasting, with the Fourier transform~\cite{oppenheim1999discrete} playing a central role in analyzing spectral structures. However, when applied to non-stationary real world time series, frequency-domain methods face two major difficulties: \emph{spectral entanglement} that hampers effective decomposition, and the computational burden of learning with complex-valued representations. \textit{Spectral entanglement} is the overlap of trend, periodic, and noise components in the frequency domain, as illustrated in Figure~\ref{fig:trend} (d). This issue arises due to two main factors. First, the finite lookback window length $L$ introduces \textit{spectral leakage}: the resolution of the Discrete Fourier Transform (DFT) is limited by $L$ according to the Nyquist theorem \cite{oppenheim1999discrete}. Any true frequency that does not align with the discrete basis causes energy to spread across all frequencies. Second, non-stationary trends, which do not exhibit periodicity, cannot be accurately represented by a small set of frequencies, as illustrated in Figure~\ref{fig:trend} (b).

The \emph{spectral entanglement} undermines the effectiveness of conventional trend-seasonality decomposition methods~\cite{cleveland1990stl}, which are typically employed to address non-stationarity in time series. For example, the moving average~\cite{wu2021autoformer, wang2024timemixer} can be viewed as applying a fixed low-pass filter in the frequency domain, but this approach is frequency-index-based and fails to resolve spectral entanglement, as shown in Figure~\ref{fig:trend} (e). Frequency-domain approaches such as Top-$K$ selection~\cite{wu2021autoformer, zhou2022fedformer} retain only dominant frequencies by amplitude, often discarding leaked or overlapping components (Figure~\ref{fig:trend} (f)). Moreover, frequency-domain learning involves complex-valued coefficients. Prior works such as FEDformer~\cite{zhou2022fedformer} and FreTS~\cite{yi2023frequency} apply specialized matrix operations to complex numbers, but these increase model complexity and hinder the integration of standard real-valued neural networks.

In response to the challenges mentioned earlier, we introduce Frequency Decomposition Network (FreDN), a novel approach that integrates learnable Frequency Disentangler and ReIm Block. At its core, FreDN uses a \textbf{Frequency Disentangler} that separates trend and seasonal components across the entire frequency spectrum, reducing spectral overlap. The disentangled trend components are processed in the time domain, where gradual variations are better captured. Seasonal components, are modeled in the frequency domain using a theory-guided \textbf{ReIm Block}, which avoids direct complex-valued computation by decomposing complex representations into real and imaginary parts with shared weights, reducing parameter cost by over 50\% without degrading performance. We further conduct a theoretical analysis comparing the gradient properties of loss functions defined in the time and frequency domains, highlighting the structured and informative nature of frequency-domain MAE. The main contributions of this work are as follows:

\begin{itemize}
    \item We highlight spectral entanglement as an overlooked issue in frequency-domain learning and proposes a simple yet effective solution to mitigate its impact through a \emph{Frequency Disentangler}.
    \item We propose a theory-guided \emph{ReIm Block} that models real and imaginary parts as dual real-valued branches. This design reduces parameter count by over 50\% without sacrificing performance, and offers a practical pathway to adapt real-valued architectures for learning in the complex plane.
    \item We present a theoretical analysis comparing the gradient propagation behaviors of MSE and MAE losses in both time and frequency domains, highlighting the structured gradients induced by frequency-domain MAE.
\end{itemize}

\section{Related Work}
\subsection{Time Series Forecasting}
Time series forecasting plays a crucial role in fields such as finance, energy, and climate science. Classical statistical models like ARMA, ARIMA, and VAR~\cite{box2015time, makridakis1997arma, watson1994vector, brockwell1991time} rely on linear assumptions and are limited in capturing nonlinear dependencies. Deep learning methods significantly improve modeling capacity, with RNNs~\cite{salinas2020deepar}, CNNs~\cite{bai2018empirical, liu2022scinet}, Transformers~\cite{wu2021autoformer, zhou2021informer, nie2023patchtst, liu2024itransformer, wen2023transformers}, and other task-specific architectures~\cite{du2020learning, du2021vtnet, du2023probabilistic, Du2023CVPR} demonstrating strong performance on complex temporal patterns. Recent efforts emphasize efficiency and fairness through compact designs~\cite{wang2024timemixer, han2024softs, lin2024sparsetsf, qiu2024tfb, qiu2025duet, qiu2025DBLoss}. Frequency-domain modeling has gained attention for its ability to capture global periodic structures and suppress noise~\cite{zhang2025timeseriesanalysisfrequency}. Approaches like FreTS, FITS, and FreDF~\cite{yi2023frequency, xu2024fits, wang2025fredf} operate directly on spectral representations, while models such as CoST and FiLM~\cite{woo2022cost, zhou2022film} incorporate frequency-domain operations into neural architectures. Broader spectral paradigms, such as Fourier Neural Operators and wavelet CNNs~\cite{li2021fourier, fang2024spiking}, demonstrate strong modeling capacity.

\subsection{Time Series Decomposition}
To handle non-stationarity, time-domain methods often decompose sequences into trend, seasonal, and noise components. Classical approaches include differencing~\cite{oppenheim1999discrete}, STL~\cite{cleveland1990stl}, SSA~\cite{golyandina2001analysis}, and EMD~\cite{huang1998empirical}. While effective, these techniques are tailored to time-domain processing and do not adapt well to frequency-domain tasks. Non-stationary sequences pose additional challenges in the frequency domain, where overlapping spectral energy from trend and periodic components causes \emph{spectral entanglement}~\cite{zhang2025timeseriesanalysisfrequency}. Heuristic solutions like Top-$K$ frequency selection~\cite{wu2021autoformer, zhou2022fedformer} are limited by spectral leakage and lack structural interpretability. Though localized spectral methods like STFT~\cite{oppenheim1999discrete} offer partial solutions, they are constrained by fixed resolution or high computational cost. Unlike our FreDN, which targets spectral entanglement via trend-seasonal separation, recent methods like FreDo~\cite{sun2022fredo}, FAITH~\cite{li2025faith}, and FreqMoE~\cite{liu2025freqmoe} either model frequencies holistically or decompose signals differently without resolving frequency overlap. To address these limitations, we propose FreDN, a learnable frequency-domain decomposition framework that disentangles trend and periodic components via adaptive frequency disentangler, while preserving compatibility with real-valued networks.
\section{Method}
\label{sec:method}
\subsection{Preliminary}
\noindent \textbf{Problem Definition.}
The input multivariate time series is $X \in \mathbb{R}^{C \times L}$, where $C$ is the number of features and $L$ is the look-back window length. The target of forecasting is $Y \in \mathbb{R}^{C \times \tau}$, where $\tau$ denotes the prediction horizon. Our forecasting model is denoted as $\hat{Y} = f_{\theta}(X)$, where $f_{\theta}$ represents the parameterized forecasting model. Moreover, $L_{\text{freq}} = \left\lfloor \frac{L}{2} \right\rfloor + 1$ and $\tau_{\text{freq}} = \left\lfloor \frac{\tau}{2} \right\rfloor + 1$ denote the frequency-domain lengths of the input and target sequences after real-valued FFT. We denote the real and imaginary parts of any complex-valued variable $\tilde{Z}$ by $\tilde{Z}_{r}$ and $\tilde{Z}_{i}$. 

\begin{figure*}[t]
  \centering
  \includegraphics[width=0.95\textwidth]{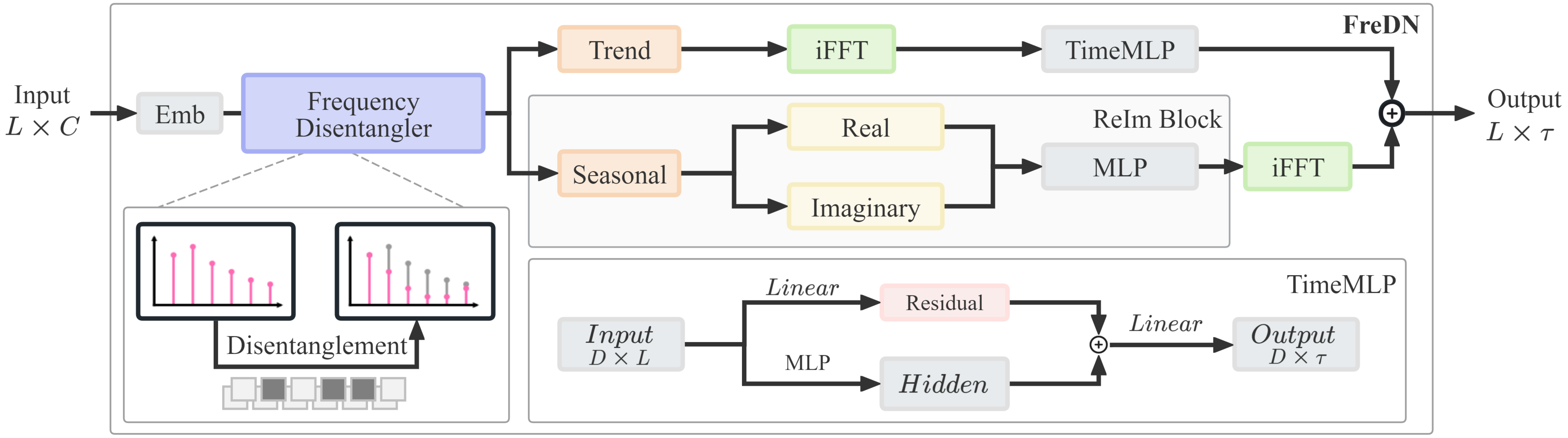}
  \caption{Overview of the proposed FreDN architecture. The input is first embedded by adding a redundant dimension, then decomposed by a learnable Frequency Disentangler into trend and seasonal components. The trend is transformed back to the time domain and processed by a TimeMLP. The seasonal real and imaginary parts are separately modeled by a shared MLP under the ReIm Block. The final prediction sums both outputs.
}
  \label{fig:model}
\end{figure*}

The overall architecture of the FreDN model is illustrated in Figure~\ref{fig:model}.

\noindent \textbf{Reversible Instance Normalization.} 
We normalize the input using Reversible Instance Normalization (RevIN)~\cite{kim2021reversible}, following the common practice in many models~\cite{nie2023patchtst, liu2024itransformer, han2024softs}. 

\noindent \textbf{Series Embedding. }
To increase the model's expressiveness, the input $X \in \mathbb{R}^{C \times L}$ is reshaped to $X \in \mathbb{R}^{C \times L \times 1}$ and projected into a $d$-dimensional embedding space: $X_{\text{emb}} = X \odot \phi_d \in \mathbb{R}^{C \times L \times d}$ with learnable $\phi_d$.

\noindent \textbf{Discrete Fourier Transformation.}
Given an input $X \in \mathbb{R}^{L}$, the DFT is defined as:
\[
\tilde{X}_k = \sum_{t=0}^{L-1} X_t \cdot e^{-j 2\pi kt / L}, \quad k = 0, 1, \ldots, L-1,
\label{eq:fft}
\]
where $j$ is the imaginary unit. The DFT is a linear transformation and can be expressed as: $\tilde{X} = F \cdot X$, where $F \in \mathbb{C}^{L \times L}$ is the Fourier matrix satisfies the unitary condition $F\cdot F^H = I$, where $F^H$ is the conjugate transpose of $F$. We use the Fast Fourier Transform (FFT) as an efficient algorithm for computing the DFT.

\subsection{Frequency Disentangler}
\label{sec:FD}
Due to the spectral entanglement across all frequencies, existing methods such as moving average and Top-$K$ selection~\cite{zhou2022fedformer, wu2021autoformer} fail to function effectively in the frequency domain. To address this issue, we design a learnable Frequency Disentangler that adaptively separates the entire spectrum into smooth trend and seasonal components, effectively alleviating spectral entanglement.

The following theorem shows that smooth trends have non-zero spectral energy at all frequencies and decay gradually with frequency index \( k \). Intuitively, smooth trend functions cannot be accurately represented by a finite set of periodic Fourier bases; their spectra typically consist of energy-concentrated low frequencies and small-magnitude high frequencies. Detailed proofs are provided in Appendix D.

\begin{theorem}[Spectral of Sobolev-Smooth Trends]
\label{thm:trend}
Let $f \in W^{m,2}([0,1])$ be a sobolev-smooth function with square-integrable $m$-th derivative for some $m \geq 1$. $\hat{f}(k)$ is the $k$-th Fourier coefficient of the periodic extension of $f$. Then there exists a constant $C > 0$ such that for all $k \in \mathbb{Z} \setminus \{0\}$,
\[
0 \leq |\hat{f}(k)| \leq \frac{C}{|k|^m}.
\]
\end{theorem}
This sobolev-smooth assumption encompasses a wide class of commonly used trend representations, including B-splines, local polynomial regression, and kernel-based smoothers, all of which produce functions within Sobolev spaces \( W^{m,2} \) for moderate \( m \). The resulting trend component is sufficiently smooth without being overly restrictive. 

Moreover, due to the finite window length used in DFT, \textit{spectral leakage} inevitably occurs, spreading the energy of seasonal component across adjacent frequencies. This leads to entanglement of trend and seasonality across the entire spectrum. To address this entanglement, we adopt a learnable decomposition strategy that assigns trend and seasonal attribution to each frequency component independently. This design allows the model to flexibly adapt to the intrinsic spectral structure of the input. 

Given the embedded input \( X_{\text{emb}} \in \mathbb{R}^{C \times L \times d} \), we apply FFT along the temporal axis and obtain:
\[
\tilde{X} = \text{FFT}(X_{\text{emb}}) \in \mathbb{C}^{C \times L_{\text{freq}} \times d}.
\]
We then introduce a learnable frequency disentangler \( M \in \mathbb{R}^{L_{\text{freq}} \times d} \), and define:
\[
\tilde{X}_{\text{trend}} = \tilde{X} \odot \sigma(M), \quad 
\tilde{X}_{\text{season}} = \tilde{X} \odot (1 - \sigma(M)),
\]
where \( \sigma \) is the sigmoid function and \( \odot \) denotes element-wise multiplication. The trend component is transformed back via inverse FFT, while the seasonal part \( \tilde{X}_{\text{season}} \) remains in the frequency domain for further modeling. The decay pattern of sobolev-smooth trend also guides the initialization of the disentangler. For instance, when \( m = 1 \), one may initialize the trend weights using \( w(k) = -\log(1 + |k|) \), which approximates the expected spectral decay of smooth trends. 

Our learnable Frequency Disentangler performs separation of trend and seasonality at each frequency, achieving fine-grained disentanglement of spectral components and yielding a smoothly varying trend spectrum as shown in Figure~\ref{fig:trend} (a) (b) alongside a decomposed, relatively stationary seasonal spectrum.

\noindent \textbf{TimeMLP. }  
We adopt a standard MLP with residual connection, where the input passes through a multilayer perceptron and a linear residual path before being projected to the target length. Specifically, TimeMLP learns smooth trend representations in the time domain. Detailed architectures are provided in Appendix B.

\subsection{ReIm Block}
\label{sec:ReIm}
Existing methods such as FreTS~\cite{yi2023frequency} and FITS~\cite{xu2024fits} commonly adopt standard complex-valued linear layers to process frequency components. However, this approach requires specially designed complex matrix multiplications, introducing additional implementation complexity and computational overhead.

A standard complex-valued linear layer takes the form:
\[
\tilde{Y} = (W_r + j W_i)(\tilde{X}_r + j \tilde{X}_i),
\]
where \( W_r \) and \( W_i \) are real-valued. To understand the role of each component, we consider the following result.

\begin{definition}[Complex Linear Projection]
Let \( \tilde{X} = \tilde{X}_r + j \tilde{X}_i \in \mathbb{C}^d \) be a complex-valued input, and let \( W = W_r + j W_i \in \mathbb{C}^{1 \times d} \) be a complex weight vector. Then the output \( \tilde{Y} = W \tilde{X} \) can be written as:
\[
\tilde{Y} = W_r \tilde{X} + j W_i \tilde{X}.
\]
Here, \( W_r \tilde{X} \) is a real-weighted projection of the input, and \( j W_i \tilde{X} \) performs an orthogonal rotation \(90^\circ\) of the same input with separate weights.
\end{definition}

\begin{table*}[t]
\centering
\setlength{\tabcolsep}{2.9pt}
\begin{tabular}{c|cc|cc|cc|cc|cc|cc|cc|cc}
\toprule
\multirow{2}{*}{Models} & \multicolumn{2}{c|}{FreDN} & \multicolumn{2}{c|}{FreDF} & \multicolumn{2}{c|}{SOFTS} & \multicolumn{2}{c|}{iTransformer} & \multicolumn{2}{c|}{TimeMixer} & \multicolumn{2}{c|}{FreTS} & \multicolumn{2}{c|}{PatchTST} & \multicolumn{2}{c}{DLinear} \\
 & \multicolumn{2}{c|}{(Ours)} & \multicolumn{2}{c|}{(2025)} & \multicolumn{2}{c|}{(2024)} & \multicolumn{2}{c|}{(2024)} & \multicolumn{2}{c|}{(2023)} & \multicolumn{2}{c|}{(2023)} & \multicolumn{2}{c|}{(2023)} & \multicolumn{2}{c}{(2023)} \\
\midrule
Metrics & MSE & MAE & MSE & MAE & MSE & MAE & MSE & MAE & MSE & MAE & MSE & MAE & MSE & MAE & MSE & MAE \\
\midrule
ETTh1 & \textbf{0.397} & \textbf{0.420} & 0.482 & 0.473 & 0.421 & 0.436 & 0.438 & 0.448 & 0.430 & 0.440 & 0.463 & 0.463 & 0.424 & 0.440 & \underline{0.416} & \underline{0.435} \\
ETTh2 & \textbf{0.330} & \textbf{0.380} & 0.371 & 0.405 & 0.361 & 0.401 & 0.377 & 0.406 & 0.359 & 0.399 & 0.462 & 0.467 & \underline{0.351} & \underline{0.391} & 0.483 & 0.472 \\
ETTm1 & \textbf{0.340} & \textbf{0.372} & 0.375 & 0.395 & 0.360 & 0.389 & 0.366 & 0.393 & 0.362 & 0.386 & 0.370 & 0.391 & 0.363 & 0.384 & \underline{0.355} & \underline{0.378} \\
ETTm2 & \textbf{0.243} & \textbf{0.304} & 0.259 & 0.318 & 0.273 & 0.325 & 0.270 & 0.329 & 0.260 & 0.317 & 0.274 & 0.333 & \underline{0.254} & \underline{0.316} & 0.255 & \underline{0.316} \\
Weather & \textbf{0.215} & \textbf{0.253} & 0.233 & 0.270 & 0.242 & 0.276 & 0.237 & 0.272 & 0.228 & \underline{0.266} & \underline{0.224} & 0.274 & 0.235 & 0.274 & 0.238 & 0.289 \\
Electricity & \textbf{0.155} & \textbf{0.247} & \underline{0.160} & \underline{0.253} & 0.161 & 0.256 & 0.161 & 0.257 & 0.165 & 0.258 & 0.166 & 0.300 & 0.164 & 0.283 & 0.164 & 0.297 \\
Traffic & 0.381 & \underline{0.261} & 0.411 & 0.290 & \textbf{0.376} & \textbf{0.260} & \underline{0.379} & 0.270 & 0.386 & 0.267 & 0.431 & 0.300 & 0.402 & 0.283 & 0.423 & 0.297 \\
\midrule
1st Count & 6 & 6 & 0 & 0 & 1 & 1 & 0 & 0 & 0 & 0 & 0 & 0 & 0 & 0 & 0 & 0 \\
2nd Count & 0 & 1 & 1 & 1 & 0 & 0 & 1 & 0 & 0 & 1 & 1 & 0 & 2 & 2 & 2 & 3 \\
\bottomrule
\end{tabular}
\caption{Multivariate forecasting results averaged over prediction horizons $\tau \in \{96, 192, 336, 720\}$. For each $\tau$, the best result is selected from lookback lengths $L \in \{96, 192, 336, 512, 720\}$ for all models. The best results are in bold, and the second best are underlined.}
\label{tab:main}
\end{table*}

This implies that the standard complex projection linearly combines the original input and its quadrature component, allowing for arbitrary complex rotations. The imaginary unit \( j \) provides a canonical mechanism for representing a phase rotation of \(90^\circ\). 

\begin{table}[t]
\centering
\setlength{\tabcolsep}{3pt}
\begin{tabular}{c|c|c|c|c|c}
\toprule
Model & Metric & ETT & Weather & Electricity & Traffic \\
\midrule
\multirow{2}{*}{FreDN} 
  & MSE & \textbf{0.328} & \textbf{0.215} & \textbf{0.155} & \textbf{0.381} \\
  & MAE & \textbf{0.369} & \textbf{0.253} & \textbf{0.247} & \textbf{0.261} \\
\midrule
\multirow{2}{*}{MovDN} 
  & MSE & 0.337 & 0.219 & 0.158 & 0.388 \\
  & MAE & 0.373 & 0.254 & 0.250 & 0.269 \\
\midrule
\multirow{2}{*}{TopKDN} 
  & MSE & 0.338 & 0.222 & \textbf{0.155} & 0.389 \\
  & MAE & 0.374 & 0.258 & 0.248 & 0.268 \\
\bottomrule
\end{tabular}
\caption{Comparison of decomposition methods. Results are averaged over prediction lengths $\tau \in \{96,192,336,720\}$, with ETT representing the average across ETTh1, ETTh2, ETTm1, and ETTm2.}
\label{tab:decomp}
\end{table}

If we set \( W_i = 0 \), the operation degenerates to \( \tilde{Y} = W_r \tilde{X} \), which corresponds to the projection strategy used in the proposed ReIm Block. Instead of relying on full complex-valued multiplications, ReIm Block adopts shared real-valued weights, significantly simplifying implementation while retaining sufficiency. The following theorem formalizes its representational capacity:
\begin{theorem}[ReIm Block]
Let \( \tilde{X} = [\tilde{X}_1, \dots, \tilde{X}_d]^\top \in \mathbb{C}^d \) be a complex-valued input vector.
\begin{itemize}
    \item \textbf{(Complex Linear Projection)} Let \( W \in \mathbb{C}^d \). Then the standard complex projection $\tilde{Y}_{\text{complex}} = W^\top \tilde{X}$ can represent any complex value.
    \item \textbf{(ReIm Block)} Let \( W_r \in \mathbb{R}^d \), the projection$
    \tilde{Y}_{\text{real}} = W_r^\top \tilde{X}$
    can represent any complex number in \( \mathbb{C} \) if and only if there exist at least two entries \( \tilde{X}_i, \tilde{X}_j \) such that their phase difference satisfies
    $\arg(\tilde{X}_i) - \arg(\tilde{X}_j) \notin \pi \mathbb{Z}.$ That is, the input contains at least two linearly independent directions in the complex plane.
\end{itemize}
\end{theorem}
Intuitively, the complex plane is a two-dimensional space, so any complex number can be represented as a linear combination of two orthogonal complex numbers (i.e., with a \(90^\circ\) phase difference). This shows ReIm retains ComplexLinear’s key feature: frequency interaction.

Given the theoretical justification above, we define the ReIm Block as: 
\[
\tilde{Y}_{\text{season}} = \text{MLP}(\tilde{X}_{\text{season}}^{(r)})+j \cdot \text{MLP}(\tilde{X}_{\text{season}}^{(i)}), 
\]
Due to the spectral complexity of real-world time series and the increasing number of frequency components with longer inputs, a real-valued projection as in ReIm Block is sufficient to approximate a wide range of complex representations through linear combinations.

Finally, we obtain the prediction by inverse transforming and combining with the trend:
\[
\hat{Y} = \text{TimeMLP}(X_{\text{trend}}) + \text{IFFT}(\tilde{Y}_{\text{season}}). 
\]
\subsection{Loss Function Analysis}
\label{sec:loss}
Here, we provide a gradient-based analysis to examine why the frequency-domain MAE loss, first introduced in FreDF~\cite{wang2025fredf}, may offer more efficient optimization behavior.
\[
\mathcal{L}_F = \frac{1}{\tau_{\text{freq}}} \| \tilde{\epsilon} \|_1.
\]
FreDF~\cite{wang2025fredf} claims that ''by transforming the label sequence into this orthogonal frequency domain, the dependence from label autocorrelation could be effectively mitigated''. This statement is statistically inaccurate. As a linear transformation, DFT preserves the correlation structure of residuals, unless the residuals are strictly independent. This misunderstands the role of orthogonality: orthogonality enables frequency separation, not decorrelation.

Despite this, the frequency-domain MAE loss is highly effective in practice, particularly when integrated into spectral learning frameworks. We now examine how different loss functions affect learning by comparing their gradient behavior with respect to $\hat{Y}$:

\textbf{Time-domain MSE:}
\[
\mathcal{L}_{\text{time-MSE}} = \frac{1}{\tau} \| \epsilon \|_2^2, \quad 
\frac{\partial \mathcal{L}_{\text{time-MSE}}}{\partial \hat{Y}} = \frac{2}{\tau} \cdot \epsilon
\]

\textbf{Frequency-domain MSE:}
\[
\mathcal{L}_{\text{freq-MSE}} = \frac{1}{\tau_{\text{freq}}} \| \tilde{\epsilon} \|_2^2, \quad 
\frac{\partial \mathcal{L}_{\text{freq-MSE}}}{\partial \hat{Y}} = \frac{2}{\tau_{\text{freq}}} \cdot \epsilon
\]

\begin{figure*}[t]
  \centering
  \includegraphics[width=0.95\textwidth]{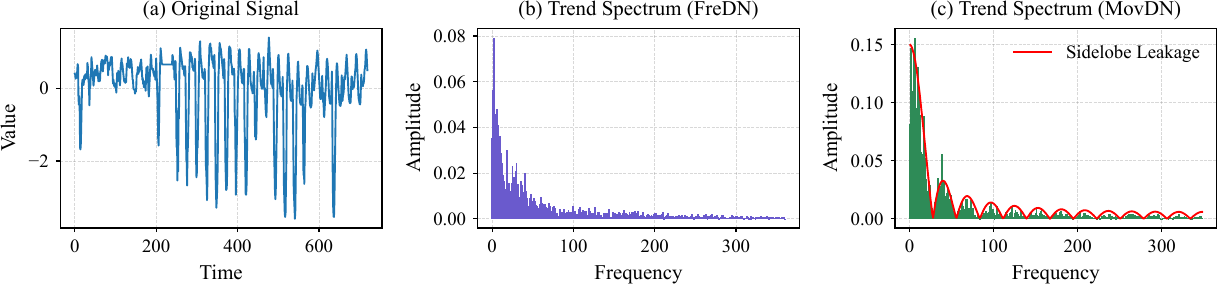}
  \caption{(a) Original signal from the ETTh1 with $L=720$; (b) Trend spectrum in our FreDN; (c) Trend spectrum learned by MovDN using moving average, the red line indicating the theoretical sidelobe leakage introduced by the moving average filter.}
  \label{fig:realdecomp}
\end{figure*}

\begin{figure*}[t]
  \centering
  \includegraphics[width=0.95\textwidth]{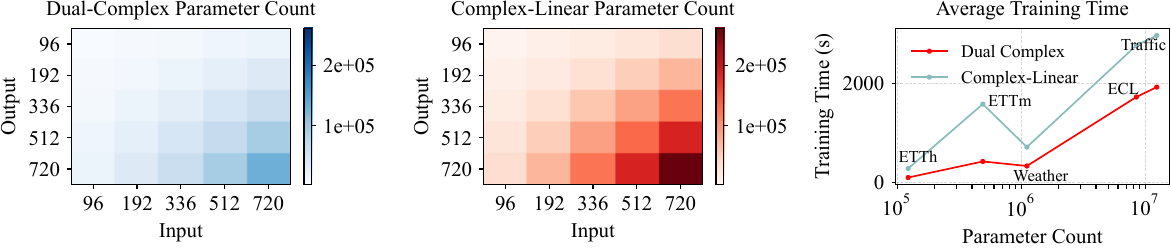}
  \caption{Efficiency of ReIm Block. (a) Parameter count of the ReIm Block under varying input/output lengths; (b) Parameter count of the Complex-Linear structure; (c) Average training time on different datasets. }
  \label{fig:efficiency}
\end{figure*}

Specifically, from the Parseval's theorem~\cite{oppenheim1999discrete}, the energy of the residual is preserved under DFT, which means $\| \epsilon \|_2^2 = \| \tilde{\epsilon} \|_2^2.$ This implies that time-MSE and frequency-MSE differ only by a constant factor in both loss magnitude and gradient.

\textbf{Time-domain MAE:}
\[
\mathcal{L}_{\text{time-MAE}} = \frac{1}{\tau} \| \epsilon \|_1, \quad 
\frac{\partial \mathcal{L}_{\text{time-MAE}}}{\partial \hat{Y}} = \frac{1}{\tau} \cdot \text{sign}(\epsilon)
\]

\textbf{Frequency-domain MAE:}
\[
\mathcal{L}_{\text{freq-MAE}} = \frac{1}{\tau_{\text{freq}}} \| \tilde{\epsilon} \|_1, \quad 
\frac{\partial \mathcal{L}_{\text{freq-MAE}}}{\partial \hat{Y}} = \frac{1}{\tau_{\text{freq}}} \cdot F^H \left( \frac{\tilde{\epsilon}}{|\tilde{\epsilon}|} \right)
\]

\begin{table}[t]
\centering
\setlength{\tabcolsep}{3pt}
\begin{tabular}{c|c|c|c|c|c}
\toprule
Structure & Metric & ETT & Weather & Electricity & Traffic \\
\midrule
    Dual & MSE & \textbf{0.328} & \textbf{0.215} & \textbf{0.155} & \textbf{0.381} \\
  Complex & MAE & \textbf{0.369} & \textbf{0.253} & \textbf{0.247} & \textbf{0.261} \\
\midrule
  Complex & MSE & 0.335 & 0.216 & 0.157 & 0.394 \\
  Linear & MAE & 0.371 & \textbf{0.253} & 0.250 & 0.270 \\
\bottomrule
\end{tabular}
\caption{Performance comparison between Dual-Complex and Complex-Linear. ETT (avg) denotes the average of ETTh1, ETTh2, ETTm1, and ETTm2. Results are averaged over prediction lengths $\tau \in \{96, 192, 336, 720\}$.}
\label{tab:complex}
\end{table}

In time-domain MAE, the gradient at each time step is a real number of unit magnitude, determined solely by the sign of the residual. This results in sparse, local signals, where each step receives an isolated update.

In contrast, frequency-domain MAE produces complex gradients of unit magnitude. The complex gradients carries a specific phase, which encodes the temporal shift (i.e., starting position) of a sinusoidal basis in the time domain, which modulates a global periodic component across the entire sequence. \emph{As a result, frequency-domain MAE introduces structured, long-range interactions into the gradient signal, enabling the model to capture global temporal patterns aligned with dominant frequencies.} 

Detailed proofs are provided in Appendix D. 

\section{Experiments}
\label{sec:Experiments}
\noindent \textbf{Datasets. }
To evaluate the performance of FreDN, we conduct extensive experiments on seven widely used long-term time series forecasting benchmarks: ETTh1, ETTh2, ETTm1, ETTm2, Weather, Electricity, and Traffic~\cite{wu2021autoformer,wang2025fredf,zhou2021informer}. Detailed dataset descriptions are provided in Appendix A.

\begin{figure*}[t]
  \centering
  \includegraphics[width=0.95\textwidth]{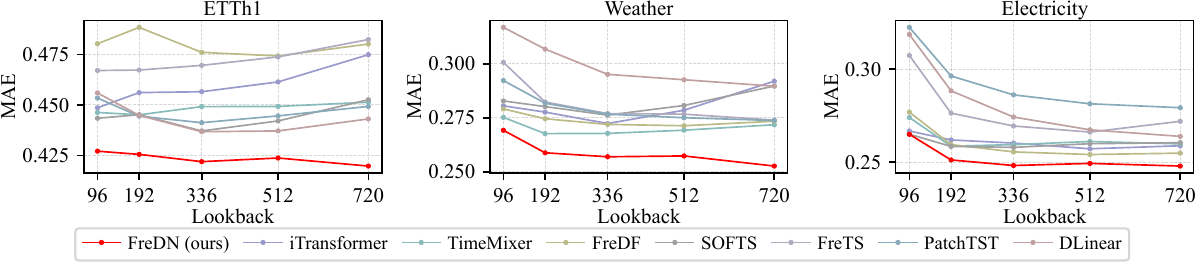}
  \caption{Effect of lookback window length on average MAE across all prediction horizons. FreDN maintains consistently superior accuracy under different lookback settings, with performance improving monotonically as the input length increases.}
  \label{fig:lookback}
\end{figure*}

\noindent \textbf{Baselines. } We compare our model against a set of representative Linear-based or MLP-based baselines, including SOFTS~\cite{han2024softs}, TimeMixer~\cite{wang2024timemixer}, DLinear~\cite{zeng2023transformers}. Two Frequency-based methods include FreTS~\cite{yi2023frequency}, FreDF~\cite{wang2025fredf}. We also consider Transformer-based methods including iTransformer~\cite{liu2024itransformer} and PatchTST~\cite{nie2023patchtst}.

\noindent \textbf{Implementation. } The baseline models are reproduced using the official training scripts provided by FreDF~\cite{wang2025fredf}. All models are optimized using the Adam optimizer~\cite{kingma2014adam}. For FreDN, we use only the frequency-domain MAE loss without the time and frequency domains fused loss structure employed by FreDF~\cite{wang2025fredf}. \emph{To ensure fairness, we follow the common practice of selecting hyperparameters based on the prediction length. For all models, the same set of hyperparameters is applied across all input lengths $\{96, 192, 336, 512, 720\}$ for a given prediction length.} All experiments are implemented in PyTorch~\cite{paszke2019pytorch} and conducted on a single NVIDIA GeForce RTX 3090 GPU with 24GB memory. More experimental details are provided in Appendix B. 
\subsection{Main results}
Table~\ref{tab:main} reports long-term forecasting results on seven benchmarks averaged over $\tau \in \{96, 192, 336, 720\}$. Each model uses default settings, and for every prediction length, the best result is selected across input lengths $L \in \{96, 192, 336, 512, 720\}$. FreDN achieves the best performance on most datasets. On average MSE, it surpasses FreDF~\cite{wang2025fredf} and FreTS~\cite{yi2023frequency} by 10\% and 14\%, and outperforms recent time-domain MLPs SOFTS~\cite{han2024softs} and TimeMixer~\cite{wang2024timemixer} by 6\%. Although our best results with lookback window selecting outperform TimeMixer++~\cite{wang2025timemixer} (as shown in the Appendix C), we exclude it for fairness, since TimeMixer++~\cite{wang2025timemixer} adopts a fixed $L=96$ and its official implementation is not publicly available. The hyperparameter sensitivity analysis are provided in Appendix C. We also report the standard deviation of FreDN performance under five runs with different random seeds in Appendix E, which exhibits that the performance is stable. We further provide the visualization analysis in Appendix F.

\subsection{Frequency Disentanglement Analysis}

The comparison of different decomposition methods is shown in Table~\ref{tab:decomp}. FreDN adopts our proposed Frequency Disentangler. MovDN applies a moving average with a fixed window size of 25, which follows the strategy used in TimeMixer~\cite{wang2024timemixer}. TopKDN selects the frequency components with the largest $K$ amplitudes as the periodic part. Detailed configurations of MovDN and TopKDN are provided in Appendix B.

In Figure~\ref{fig:realdecomp}, we present the decomposition of real world sequence. Figure~\ref{fig:realdecomp} (b) shows the trend spectrum obtained using the Frequency Disentangler. The trend obtained by moving average in Figure~\ref{fig:realdecomp}(c) exhibits significant \emph{sidelobe leakage}, characterized by a dominant main lobe accompanied by multiple high-energy sidelobes. \emph{This structure does not reflect the true spectral property of the trend component, but rather arises as a systematic artifact introduced by the decomposition method.}
 This is because the finite moving window causes spectral leakage in the frequency domain, which is a direct consequence of the Nyquist theorem~\cite{oppenheim1999discrete}. Specifically, the moving average decomposition is equivalent to a filter with the following frequency response:
\[
H(f) = \frac{1}{k} \cdot \frac{\sin(\pi f k)}{\sin(\pi f)} \cdot e^{-j \pi f (k-1)}
\]
where \( k \) represents the window length of the moving average, and \(\frac{\sin(\pi f k)}{\sin(\pi f)}\) describes the low-pass filter characteristics with sidelobes, illustrated by the red curve in Figure~\ref{fig:realdecomp} (c). 
This phenomenon further underscores the inadequacy of moving average techniques for frequency-based processing and supports the necessity of our Frequency Disentangler to decompose directly in the frequency domain.

\subsection{Efficiency of ReIm Block}

In Section \textbf{ReIm Block}, we provided a theoretical justification for the ReIm Block. To further demonstrate its practical efficiency, we compare the parameter count and average training time of the proposed ReIm Block with a Complex-Linear structure.  The Complex-Linear structure replaces the linear projections in MLP with complex-valued linears, where the detailed architecture is described in Appendix B. Specifically, to implement a Complex-Linear, it requires two sets of parameters for real and imaginary parts, and follows the complex matrix rule~\cite{golub2013matrix}: 
\[
\tilde{W} \cdot \tilde{X} = \tilde{W}^{(r)} \tilde{X}^{(r)} - \tilde{W}^{(i)} \tilde{X}^{(i)} + j \cdot (\tilde{W}^{(r)} \tilde{X}^{(i)} + \tilde{W}^{(i)} \tilde{X}^{(r)}).
\]

As shown in Figure~\ref{fig:efficiency}, FreDN with ReIm Block reduces the parameter count by approximately 50\% under the same configuration. Furthermore, it achieves significant training speedup, reducing average runtime by up to 75\% on the ETTm dataset. Full results are provided in Appendix C.

\noindent \textbf{Comparison of different complex-valued structures.}
We compare the average forecasting performance of the proposed ReIm Block with Complex-Linear structure in Table~\ref{tab:complex}. Despite achieving a significant reduction in model size and training cost as demonstrated in Figure~\ref{fig:efficiency}, the performance of the ReIm Block is not compromised, and even slightly outperforms the Complex-Linear structure. This may be attributed to the structural limitations of complex-valued representations, such as the lack of activation functions or layer normalization in complex form, highlighting the structural advantage of the ReIm Block design.

\noindent \textbf{Effect of lookback window length.} 
Although longer input sequences contain richer temporal information, lots of models tend to suffer from performance degradation as the lookback length increases. This is likely due to the weaker correlation between distant inputs and future targets, which poses challenges to the model's ability to capture global dependencies. We explore how varying the lookback length $L \in \{96, 192, 336, 512, 720\}$ affects forecasting accuracy across all datasets. Figure~\ref{fig:lookback} presents the results, which shows FreDN achieving stable performance improvements as $L$ increases. The performance improvement of FreDN over other models becomes more pronounced with longer inputs, indicating that FreDN is better equipped to exploit the additional information provided by extended lookback windows.

\section{Conclusion}
We present a novel frequency-domain perspective for non-stationary time series forecasting, focusing on spectral entanglement issue. To address this, we propose \textbf{FreDN}, which leverages a learnable Frequency Disentangler to disentangle trends and seasonalities directly in the frequency domain. We also design the ReIm Block block with a single real-valued MLP to model complex spectra efficiently. We theoretically justify the feasibility of ReIm Block and analyze the frequency-domain MAE loss through a gradient propagation view, showing its advantage in capturing global dependencies. ReIm Block provides an insightful perspective for frequency-domain modeling, which may inspire future architectural innovations. Empirically, FreDN ranks first in 26 out of 28 MAE tasks across four prediction lengths and seven baselines (see Appendix C).

\bibliography{aaai2026}

@book{brockwell1991time,
  title={Time series: Theory and methods},
  author={Brockwell, Peter J},
  year={1991},
  publisher={Springer-Verlag}
}

@inproceedings{zeng2023transformers,
  title={Are transformers effective for time series forecasting?},
  author={Zeng, Ailing and Chen, Muxi and Zhang, Lei and Xu, Qiang},
  booktitle={Proceedings of the AAAI conference on artificial intelligence},
  volume={37},
  number={9},
  pages={11121--11128},
  year={2023}
}

@inproceedings{lin2024sparsetsf,
title={Sparse{TSF}: Modeling Long-term Time Series Forecasting with *1k* Parameters},
author={Shengsheng Lin and Weiwei Lin and Wentai Wu and Haojun Chen and Junjie Yang},
booktitle={Forty-first International Conference on Machine Learning},
year={2024},
url={https://openreview.net/forum?id=54NSHO0lFe}
}

@inproceedings{han2024softs,
title={{SOFTS}: Efficient Multivariate Time Series Forecasting with Series-Core Fusion},
author={Lu Han and Xu-Yang Chen and Han-Jia Ye and De-Chuan Zhan},
booktitle={The Thirty-eighth Annual Conference on Neural Information Processing Systems},
year={2024},
url={https://openreview.net/forum?id=89AUi5L1uA}
}

@inproceedings{wang2024timemixer,
title={TimeMixer: Decomposable Multiscale Mixing for Time Series Forecasting},
author={Shiyu Wang and Haixu Wu and Xiaoming Shi and Tengge Hu and Huakun Luo and Lintao Ma and James Y. Zhang and JUN ZHOU},
booktitle={The Twelfth International Conference on Learning Representations},
year={2024},
url={https://openreview.net/forum?id=7oLshfEIC2}
}

@inproceedings{wang2025timemixer,
title={TimeMixer++: A General Time Series Pattern Machine for Universal Predictive Analysis},
author={Shiyu Wang and Jiawei LI and Xiaoming Shi and Zhou Ye and Baichuan Mo and Wenze Lin and Ju Shengtong and Zhixuan Chu and Ming Jin},
booktitle={The Thirteenth International Conference on Learning Representations},
year={2025},
url={https://openreview.net/forum?id=1CLzLXSFNn}
}

@Article{salinas2020deepar,
  author    = {Salinas, David and Flunkert, Valentin and Gasthaus, Jan and Januschowski, Tim},
  journal   = {International Journal of Forecasting},
  title     = {DeepAR: Probabilistic forecasting with autoregressive recurrent networks},
  year      = {2020},
  number    = {3},
  pages     = {1181--1191},
  volume    = {36},
  publisher = {Elsevier},
}

@article{liu2022scinet,
  title={Scinet: Time series modeling and forecasting with sample convolution and interaction},
  author={Liu, Minhao and Zeng, Ailing and Chen, Muxi and Xu, Zhijian and Lai, Qiuxia and Ma, Lingna and Xu, Qiang},
  journal={Advances in Neural Information Processing Systems},
  volume={35},
  pages={5816--5828},
  year={2022}
}

@inproceedings{wen2023transformers,
  title={Transformers in time series: a survey},
  author={Wen, Qingsong and Zhou, Tian and Zhang, Chaoli and Chen, Weiqi and Ma, Ziqing and Yan, Junchi and Sun, Liang},
  booktitle={Proceedings of the Thirty-Second International Joint Conference on Artificial Intelligence},
  pages={6778--6786},
  year={2023}
}

@inproceedings{zhou2021informer,
  title={Informer: Beyond efficient transformer for long sequence time-series forecasting},
  author={Zhou, Haoyi and Zhang, Shanghang and Peng, Jieqi and Zhang, Shuai and Li, Jianxin and Xiong, Hui and Zhang, Wancai},
  booktitle={Proceedings of the AAAI conference on artificial intelligence},
  volume={35},
  number={12},
  pages={11106--11115},
  year={2021}
}

@article{wu2021autoformer,
  title={Autoformer: Decomposition transformers with auto-correlation for long-term series forecasting},
  author={Wu, Haixu and Xu, Jiehui and Wang, Jianmin and Long, Mingsheng},
  journal={Advances in neural information processing systems},
  volume={34},
  pages={22419--22430},
  year={2021}
}

@inProceedings{zhou2022fedformer,
  author       = {Zhou, Tian and Ma, Ziqing and Wen, Qingsong and Wang, Xue and Sun, Liang and Jin, Rong},
  booktitle    = {International Conference on Machine Learning},
  title        = {Fedformer: Frequency enhanced decomposed transformer for long-term series forecasting},
  year         = {2022},
  organization = {PMLR},
  pages        = {27268--27286},
}

@InProceedings{nie2023patchtst,
  author    = {Nie, Yuqi and H. Nguyen, Nam and Sinthong, Phanwadee and Kalagnanam, Jayant},
  booktitle = {International Conference on Learning Representations},
  title     = {A Time Series is Worth 64 Words: Long-term Forecasting with Transformers},
  year      = {2023},
}

@InProceedings{liu2024itransformer,
  author    = {Liu, Yong and Hu, Tengge and Zhang, Haoran and Wu, Haixu and Wang, Shiyu and Ma, Lintao and Long, Mingsheng},
  booktitle = {The Twelfth International Conference on Learning Representations},
  title     = {iTransformer: Inverted Transformers Are Effective for Time Series Forecasting},
  year      = {2024},
}

@InProceedings{kim2021reversible,
  author    = {Kim, Taesung and Kim, Jinhee and Tae, Yunwon and Park, Cheonbok and Choi, Jang-Ho and Choo, Jaegul},
  booktitle = {International Conference on Learning Representations},
  title     = {Reversible instance normalization for accurate time-series forecasting against distribution shift},
  year      = {2021},
}

@article{paszke2019pytorch,
  title={Pytorch: An imperative style, high-performance deep learning library},
  author={Paszke, Adam and Gross, Sam and Massa, Francisco and Lerer, Adam and Bradbury, James and Chanan, Gregory and Killeen, Trevor and Lin, Zeming and Gimelshein, Natalia and Antiga, Luca and others},
  journal={Advances in neural information processing systems},
  volume={32},
  year={2019}
}

@InProceedings{kingma2014adam,
  author    = {Kingma, Diederik P and Ba, Jimmy},
  booktitle = {International conference on machine learning},
  title     = {Adam: A method for stochastic optimization},
  year      = {2015},
}

@article{yi2023frequency,
  title={Frequency-domain mlps are more effective learners in time series forecasting},
  author={Yi, Kun and Zhang, Qi and Fan, Wei and Wang, Shoujin and Wang, Pengyang and He, Hui and An, Ning and Lian, Defu and Cao, Longbing and Niu, Zhendong},
  journal={Advances in Neural Information Processing Systems},
  volume={36},
  pages={76656--76679},
  year={2023}
}

@inproceedings{xu2024fits,
title={{FITS}: Modeling Time Series with \$10k\$ Parameters},
author={Zhijian Xu and Ailing Zeng and Qiang Xu},
booktitle={The Twelfth International Conference on Learning Representations},
year={2024},
url={https://openreview.net/forum?id=bWcnvZ3qMb}
}

@inproceedings{wang2025fredf,
title={Fre{DF}: Learning to Forecast in the Frequency Domain},
author={Hao Wang and Lichen Pan and Yuan Shen and Zhichao Chen and Degui Yang and Yifei Yang and Sen Zhang and Xinggao Liu and Haoxuan Li and Dacheng Tao},
booktitle={The Thirteenth International Conference on Learning Representations},
year={2025},
url={https://openreview.net/forum?id=4A9IdSa1ul}
}

@book{oppenheim1999discrete,
  title={Discrete-time signal processing},
  author={Oppenheim, Alan V},
  year={1999},
  publisher={Pearson Education India}
}

@article{cleveland1990stl,
  title={STL: A seasonal-trend decomposition},
  author={Cleveland, Robert B and Cleveland, William S and McRae, Jean E and Terpenning, Irma and others},
  journal={J. off. Stat},
  volume={6},
  number={1},
  pages={3--73},
  year={1990}
}

@misc{zhang2025timeseriesanalysisfrequency,
      title={Time Series Analysis in Frequency Domain: A Survey of Open Challenges, Opportunities and Benchmarks}, 
      author={Qianru Zhang and Yuting Sun and Honggang Wen and Peng Yang and Xinzhu Li and Ming Li and Kwok-Yan Lam and Siu-Ming Yiu and Hongzhi Yin},
      year={2025},
      eprint={2504.07099},
      archivePrefix={arXiv},
      primaryClass={cs.CE},
      url={https://arxiv.org/abs/2504.07099}, 
}

@misc{li2021fourier,
      title={Fourier Neural Operator for Parametric Partial Differential Equations}, 
      author={Zongyi Li and Nikola Kovachki and Kamyar Azizzadenesheli and Burigede Liu and Kaushik Bhattacharya and Andrew Stuart and Anima Anandkumar},
      year={2021},
      eprint={2010.08895},
      archivePrefix={arXiv},
      primaryClass={cs.LG},
      url={https://arxiv.org/abs/2010.08895}, 
}

@inproceedings{woo2022cost,
title={Co{ST}: Contrastive Learning of Disentangled Seasonal-Trend Representations for Time Series Forecasting},
author={Gerald Woo and Chenghao Liu and Doyen Sahoo and Akshat Kumar and Steven Hoi},
booktitle={International Conference on Learning Representations},
year={2022},
url={https://openreview.net/forum?id=PilZY3omXV2}
}

@inproceedings{zhou2022film,
  author={Tian Zhou and Ziqing Ma and Xue Wang and Qingsong Wen and Liang Sun and Tao Yao and Wotao Yin and Rong Jin},
  title={FiLM: Frequency improved Legendre Memory Model for Long-term Time Series Forecasting},
  year={2022},
  cdate={1640995200000},
  url={http://papers.nips.cc/paper_files/paper/2022/hash/524ef58c2bd075775861234266e5e020-Abstract-Conference.html},
  booktitle={NeurIPS},
}

@book{golyandina2001analysis,
  title={Analysis of Time Series Structure: SSA and Related Techniques},
  author={Golyandina, Nina and Nekrutkin, Vladimir and Zhigljavsky, Anatoly},
  year={2001},
  publisher={Chapman and Hall/CRC}
}

@article{huang1998empirical,
  title={The empirical mode decomposition and the Hilbert spectrum for nonlinear and non-stationary time series analysis},
  author={Huang, Norden E and Shen, Zheng and Long, Steven R and Wu, Manli C and Shih, Hsing H and Zheng, Quanan and Yen, Nai-Chyuan and Tung, Chi Chao and Liu, Henry H},
  journal={Proceedings of the Royal Society of London. Series A: mathematical, physical and engineering sciences},
  volume={454},
  number={1971},
  pages={903--995},
  year={1998},
  publisher={The Royal Society}
}

@misc{bai2018empirical,
      title={An Empirical Evaluation of Generic Convolutional and Recurrent Networks for Sequence Modeling}, 
      author={Shaojie Bai and J. Zico Kolter and Vladlen Koltun},
      year={2018},
      eprint={1803.01271},
      archivePrefix={arXiv},
      primaryClass={cs.LG},
      url={https://arxiv.org/abs/1803.01271}, 
}

@book{box2015time,
  title={Time series analysis: forecasting and control},
  author={Box, George EP and Jenkins, Gwilym M and Reinsel, Gregory C and Ljung, Greta M},
  year={2015},
  publisher={John Wiley \& Sons}
}

@article{makridakis1997arma,
  title={ARMA models and the Box--Jenkins methodology},
  author={Makridakis, Spyros and Hibon, Michele},
  journal={Journal of forecasting},
  volume={16},
  number={3},
  pages={147--163},
  year={1997},
  publisher={Wiley Online Library}
}

@article{watson1994vector,
  title={Vector autoregressions and cointegration},
  author={Watson, Mark W},
  journal={Handbook of econometrics},
  volume={4},
  pages={2843--2915},
  year={1994},
  publisher={Elsevier}
}

@book{golub2013matrix,
  title={Matrix computations},
  author={Golub, Gene H and Van Loan, Charles F},
  year={2013},
  publisher={JHU press}
}

@inproceedings{fang2024spiking,
  title={Spiking wavelet transformer},
  author={Fang, Yuetong and Wang, Ziqing and Zhang, Lingfeng and Cao, Jiahang and Chen, Honglei and Xu, Renjing},
  booktitle={European conference on computer vision},
  pages={19--37},
  year={2024},
  organization={Springer}
}

@misc{liu2025freqmoe,
      title={FreqMoE: Enhancing Time Series Forecasting through Frequency Decomposition Mixture of Experts}, 
      author={Ziqi Liu},
      year={2025},
      eprint={2501.15125},
      archivePrefix={arXiv},
      primaryClass={cs.LG},
      url={https://arxiv.org/abs/2501.15125}, 
}

@misc{sun2022fredo,
      title={FreDo: Frequency Domain-based Long-Term Time Series Forecasting}, 
      author={Fan-Keng Sun and Duane S. Boning},
      year={2022},
      eprint={2205.12301},
      archivePrefix={arXiv},
      primaryClass={cs.LG},
      url={https://arxiv.org/abs/2205.12301}, 
}

@InProceedings{Du2023CVPR,
    author    = {Du, Heming and Li, Lincheng and Huang, Zi and Yu, Xin},
    title     = {Object-Goal Visual Navigation via Effective Exploration of Relations Among Historical Navigation States},
    booktitle = {Proceedings of the IEEE/CVF Conference on Computer Vision and Pattern Recognition (CVPR)},
    month     = {June},
    year      = {2023},
    pages     = {2563-2573}
}

@article{li2025faith,
  title={FAITH: Frequency-domain Attention In Two Horizons for time series forecasting},
  author={Li, Ruiqi and Jiang, Maowei and Liu, Quangao and Wang, Kai and Feng, Kaiduo and Sun, Yue and Zhou, Xiufang},
  journal={Knowledge-Based Systems},
  volume={309},
  pages={112790},
  year={2025},
  publisher={Elsevier}
}

@inproceedings{qiu2024tfb,
title   = {{TFB}: Towards Comprehensive and Fair Benchmarking of Time Series Forecasting Methods},
author  = {Xiangfei Qiu and Jilin Hu and Lekui Zhou and Xingjian Wu and Junyang Du and Buang Zhang and Chenjuan Guo and Aoying Zhou and Christian S. Jensen and Zhenli Sheng and Bin Yang},
booktitle = {Proc. {VLDB} Endow.},
pages   = {2363--2377},
year    = {2024}
}

@inproceedings{qiu2025duet,
title   = {{DUET}: Dual Clustering Enhanced Multivariate Time Series Forecasting},
author  = {Xiangfei Qiu and Xingjian Wu and Yan Lin and Chenjuan Guo and Jilin Hu and Bin Yang},
booktitle = {SIGKDD},
pages     = {1185-1196},
year    = {2025}
}

@inproceedings{qiu2025DBLoss,
title   = {DBLoss: Decomposition-based Loss Function for Time Series Forecasting},
author  = {Xiangfei Qiu and Xingjian Wu and Hanyin Cheng and Xvyuan Liu and Chenjuan Guo and Jilin Hu and Bin Yang},
booktitle = {NeurIPS},
year    = {2025}
}

@InProceedings{du2020learning,
author="Du, Heming
and Yu, Xin
and Zheng, Liang",
editor="Vedaldi, Andrea
and Bischof, Horst
and Brox, Thomas
and Frahm, Jan-Michael",
title="Learning Object Relation Graph and Tentative Policy for Visual Navigation",
booktitle="Computer Vision -- ECCV 2020",
year="2020",
publisher="Springer International Publishing",
address="Cham",
pages="19--34",
isbn="978-3-030-58571-6"
}

@misc{du2021vtnet,
      title={VTNet: Visual Transformer Network for Object Goal Navigation}, 
      author={Heming Du and Xin Yu and Liang Zheng},
      year={2021},
      eprint={2105.09447},
      archivePrefix={arXiv},
      primaryClass={cs.CV},
      url={https://arxiv.org/abs/2105.09447}, 
}

@article{du2023probabilistic,
author = {Du, Heming and Du, Shouguo and Li, Wen},
title = {Probabilistic time series forecasting with deep non-linear state space models},
journal = {CAAI Transactions on Intelligence Technology},
volume = {8},
number = {1},
pages = {3-13},
doi = {https://doi.org/10.1049/cit2.12085},
url = {https://ietresearch.onlinelibrary.wiley.com/doi/abs/10.1049/cit2.12085},
eprint = {https://ietresearch.onlinelibrary.wiley.com/doi/pdf/10.1049/cit2.12085},
year = {2023}
}

@misc{trabelsi2018deep,
      title={Deep Complex Networks}, 
      author={Chiheb Trabelsi and Olexa Bilaniuk and Ying Zhang and Dmitriy Serdyuk and Sandeep Subramanian and João Felipe Santos and Soroush Mehri and Negar Rostamzadeh and Yoshua Bengio and Christopher J Pal},
      year={2018},
      eprint={1705.09792},
      archivePrefix={arXiv},
      primaryClass={cs.NE},
      url={https://arxiv.org/abs/1705.09792}, 
}

@book{hoffman1971linear,
  title={Linear algebra},
  author={Hoffmann, Kenneth and Kunze, Ray Alden},
  year={1971},
  publisher={Prentice-Hall Hoboken, NJ}
}
\clearpage
\appendix

\section{Datasets Description}
We conduct experiments on 7 public real-world datasets to evaluate the performance of our proposed FreDN model. The details of these datasets are listed in Table \ref{tab:dataset}.

\paragraph{ETT}\cite{zhou2021informer} \footnote{\url{https://github.com/zhouhaoyi/ETDataset}} It contains 7 factors of electricity transformer from July 2016 to July 2018, and is collected from two different electric transformers labeled with 1 and 2. ETTh1 and ETTh2 are recorded every hour, and ETTm1 and ETTm2 are recorded every 15 minutes. 

\paragraph{Weather} \cite{wu2021autoformer} \footnote{\url{https://www.bgc-jena.mpg.de/wetter/}} It collects 21 meteorological indicators, such as humidity and air temperature, from the Weather Station of the Max Planck Biogeochemistry Institute in Germany in 2020. The data sampling interval is every 10 minutes. 

\paragraph{ECL} \cite{wu2021autoformer} \footnote{\url{https://archive.ics.uci.edu/ml/datasets/ElectricityLoadDiagrams20112014}} It contains electricity consumption of 321 clients for long-term forecasting. It is collected since 01/01/2011. The data sampling interval is every 15 minutes.

\paragraph{Traffic} \cite{wu2021autoformer} \footnote{\url{http://pems.dot.ca.gov}} It contains hourly traffic data from 862 car lanes for long-term forecasting settings. It is collected from January 2015 to December 2016 with a sampling interval of every 1 hour.
\begin{table*}[htbp]
  \centering
  \small
  \caption{Detailed dataset descriptions. \textit{Channels} denotes the variate number in each dataset. \textit{Dataset Size} denotes the total number of time points in (Train, Validation, Test) split respectively. \textit{Prediction Horizon} denotes the forcasting length and four prediction settings are included in each dataset. \textit{Frequency} denotes the sampling interval of time points.}
  \resizebox{0.88\textwidth}{!}{
    \begin{tabular}{c|c|c|c|c}
    \toprule
    Dataset & Channels   & Prediction Horizon & Dataset Size & Frequency \\
    \midrule
    ETTh1 & 7     & {96, 192, 336, 720} & (8545, 2881, 2881) & Hourly \\
    \midrule
    \multicolumn{1}{c|}{ETTh2} & 7     & {96, 192, 336, 720} & (8545, 2881, 2881) & Hourly \\
    \midrule
    \multicolumn{1}{c|}{ETTm1} & 7     & {96, 192, 336, 720} & (34665, 11521, 11521) & 15min \\
    \midrule
    ETTm2 & 7     & {96, 192, 336, 720} & (34665, 11521, 11521) & 15min \\
    \midrule
    Weather & 21    & {96, 192, 336, 720} & (36792, 5271, 10540) & 10min \\
    \midrule
    ECL   & 321   & {96, 192, 336, 720} & (18317, 2633, 5261) & Hourly \\
    \midrule
    Traffic & 862   & {96, 192, 336, 720} & (12185, 1757, 3509) & Hourly \\
    \bottomrule
    \end{tabular}%
  }
  \label{tab:dataset}%
\end{table*}%

\section{Implement Details}

\subsection{Overall Architecture}

The forward inference procedure of \textbf{FreDN} is outlined in Algorithm~\ref{alg:FreDN}, where a multivariate historical input $\boldsymbol{X} \in \mathbb{R}^{L \times C}$ is normalized, embedded, decomposed in the frequency domain, and predicted via residual MLPs. Here, $L$ and $\tau$ denote the look-back and forecasting lengths, $C$ is the number of channels, and $d$ is the embedding dimension. Frequency-domain lengths are denoted as $L_{\text{freq}} = \lfloor L/2 \rfloor + 1$ and $\tau_{\text{freq}} = \lfloor \tau/2 \rfloor + 1$.

The decomposition is performed via a learnable Frequency Disentangler (FreD) in the frequency domain, separating the signal into a trend and a seasonal component. The trend component is modeled in the time domain, while the seasonal component is processed in the frequency domain through a ReIm Block. Both branches utilize a residual MLP structure defined as:
\[
Z_{\text{out}} = \mathrm{Linear}_{\text{out}}\left( \mathrm{MLP}(Z_{\text{in}}) + \mathrm{Linear}_{\text{res}}(Z_{\text{in}}) \right),
\]
where $Z_{\text{in}} \in \mathbb{R}^{C \times d \times l_{\text{in}}}$ and $Z_{\text{out}} \in \mathbb{R}^{C \times d \times l_{\text{out}}}$ with $l_{\text{in}}, l_{\text{out}} \in \{L, \tau\}$ or $\{L_{\text{freq}}, \tau_{\text{freq}}\}$ depending on the domain.

\begin{algorithm}[htbp]
\caption{Forward Inference of FreDN}
\label{alg:FreDN}
\begin{flushleft}
\textbf{Input:} Historical sequence $X \in \mathbb{R}^{L \times C}$

\textbf{Output:} Forecast result $\hat{Y} \in \mathbb{R}^{\tau \times C}$

1. $X_{\text{norm}} = \operatorname{RevIN}(X)$ \\
2. $X_{\text{emb}} = X_{\text{norm}}^\top \odot \phi_d$  \\
3. $\tilde{X} = \operatorname{FFT}(X_{\text{emb}}, \text{dim}=1)$ \\
4. $[\tilde{X}_{\text{season}}, \tilde{X}_{\text{trend}}] = \operatorname{FreD}(\tilde{X})$ \\
5. $X_{\text{trend}} = \operatorname{IFFT}(\tilde{X}_{\text{trend}}, \text{dim}=1)$ \\
6. $[\tilde{X}^{(r)}_{\text{season}}, \tilde{X}^{(i)}_{\text{season}}] = \text{ReIm-Split}(\tilde{X}_{\text{season}})$ \\
7. $\tilde{X}^{(r)}_{\tau} = \operatorname{MLP}(\tilde{X}^{(r)}_{\text{season}})$ \\
8. $\tilde{X}^{(i)}_{\tau} = \operatorname{MLP}(\tilde{X}^{(i)}_{\text{season}})$ \\
9. $\tilde{X}_{\text{season}, \tau} = \tilde{X}^{(r)} + j \cdot \tilde{X}^{(i)}$ \\
10. $X_{\text{season}, \tau} = \operatorname{IFFT}(\tilde{X}_{\text{season}, \tau}, \text{dim}=2)$ \\
11. $\hat{Y}_{\text{season}} = \operatorname{Linear}(X_{\text{season}, \tau}^\top)$ \\
12. $X_{\text{trend}, \tau} = \operatorname{TimeMLP}(X_{\text{trend}})$ \\
13. $\hat{Y}_{\text{trend}} = \operatorname{Linear}(X_{\text{trend}, \tau}^\top)$ \\
14. $\hat{Y} = \operatorname{RevIN}^{-1}(\hat{Y}_{\text{season}} + \hat{Y}_{\text{trend}})$
\end{flushleft}
\end{algorithm}

\subsection{Training and Test Details}
Following the protocol proposed in iTransformer~\cite{liu2024itransformer}, we split all datasets into training, validation, and testing subsets strictly based on chronological order to avoid information leakage. The ETT datasets are divided in a 6:2:2 ratio, while all other datasets follow a 7:1:2 split. We set the forecasting length to $\tau \in \{96, 192, 336, 720\}$ for all models and datasets to ensure comparability. Instead of using a fixed look-back window size, we vary the historical window length $L \in \{96, 192, 336, 512, 720\}$, and report the best performance achieved over different values of $L$ for each forecasting horizon. This design allows us to better evaluate the ability of models to utilize long-range temporal information. 

To determine the optimal hyperparameter configuration for our model, we perform a grid search across a predefined small range of values. We consider batch size in the set $\{8, 16, 32\}$, and the embedding size is selected from $\{4, 8, 16\}$. The hidden lyaers in ResMLP is chosen from $\{1, 2, 3, 4\}$, and the hidden size is in $\{16, 32, 64, 128, 256, 512, 720\}$. The dropout rate is sampled from $\{0, 0.1, 0.2, 0.3\}$, while the learning rate is drawn from $\{0.0001, 0.001, 0.002, 0.01\}$. 

For simplicity and in light of FreDN's rapid convergence, we fix the number of training epochs to 20 for all datasets. To prevent overfitting and maintain model robustness, we monitor validation loss during training and apply early stopping with a maximum tolerance of 5 non-improving epochs. For learning rate scheduling, we adopt a \textit{cosine} annealing strategy for datasets with larger scale and higher variability, such as Electricity and Traffic, while using a \textit{typ1} decay scheme for the remaining datasets. We adopt Mean Squared Error (MSE) and Mean Absolute Error (MAE) as the evaluation metrics, in line with standard practice in the field. 

All experiments are implemented in PyTorch~\cite{paszke2019pytorch} and conducted on a single NVIDIA GeForce RTX 3090 GPU with 24GB memory. The average training time is shown in Figure~\ref{fig:Efficiencyfull} The ADAM optimizer~\cite{kingma2014adam} is used across all experiments. We use a fixed random seed (e.g., 2025) for reproducibility unless otherwise specified.

For our proposed FreDN model, we employ the frequency-domain MAE loss introduced in FreDF~\cite{wang2025fredf}, which is defined as
\[
\mathcal{L}_{\text{freq-MAE}} = \frac{1}{\tau_{\text{freq}}} \left\| \tilde{\epsilon} \right\|_1
\]
where $\tilde{\epsilon}=F\cdot \epsilon$ denotes the residual in the frequency domain, $F$ is the DFT transform matrix, and $\tau_{\text{freq}}$ is the forcasting length in the frequency domain. All baseline models are trained using their official implementations and default loss functions. 

We adopt two widely-used evaluation metrics to assess forecasting performance: Mean Squared Error (MSE) and Mean Absolute Error (MAE). Given the ground truth values \( y_i \) and the predicted values \( \hat{y}_i \), both metrics are computed as follows:
\[
\mathrm{MSE} = \frac{1}{N} \sum_{i=1}^N (y_i - \hat{y}_i)^2, \quad
\mathrm{MAE} = \frac{1}{N} \sum_{i=1}^N |y_i - \hat{y}_i|,
\]
where \( N \) denotes the total number of predicted time points. MSE penalizes larger errors more heavily due to the squared term, while MAE provides a more robust and interpretable measure of average absolute deviation. These two metrics are complementary and jointly reflect the accuracy and stability of model predictions.

\subsection{RevIN}
Reversible Instance Normalization (RevIN)~\cite{kim2021reversible} is a commonly adopted technique to reduce distributional discrepancies between training and testing data. It removes per-instance temporal trends before modeling and restores them after prediction. Following prior works~\cite{liu2024itransformer,wang2024timemixer}, we adopt RevIN in our model. The procedure involves three steps:

\begin{enumerate}
    \item \textbf{Computing per-variable statistics.} The mean and standard deviation over the look-back window for each variable $\boldsymbol{x}_{t-L+1:t,n} = (x_{t-L+1,n}, \ldots, x_{t,n})^\top$, where $n = 1, \ldots, N$, are given by:
    \[
    \left\{
    \begin{array}{ll}
    \mu_{t,n} = \dfrac{1}{L} \sum\limits_{t=1}^{i} x_{t-i+1,n} \\[1ex]
    \sigma_{t,n} = \left( \dfrac{1}{L} \sum\limits_{t=1}^{i} (x_{t-i+1,n} - \mu_{t,n})^2 \right)^{\frac{1}{2}}
    \end{array}
    \right.
    \]
    
    \item \textbf{Instance normalization.} The input sequence is normalized using:
    \[
    \tilde{\boldsymbol{x}}_{t-L+1:t,n}^{(0)} = 
    \underbrace{
    \gamma_n \left( 
        \dfrac{\boldsymbol{x}_{t-L+1:t,n} - \mu_{t,n}}{\sqrt{\sigma_{t,n}^2 + \epsilon}} 
    \right)
    }_{\text{instance normalization}}
    + \beta_n
    \]
    where $\gamma_n$ and $\beta_n$ are learnable affine parameters, and $\epsilon$ is a small constant for numerical stability.

    \item \textbf{Instance denormalization.} For the model output $\tilde{\boldsymbol{y}}_{t+1:t+H,n}$, we apply:
    \[
    \hat{\boldsymbol{y}}_{t+1:t+H,n} = 
    \left( 
        \sqrt{\sigma_{t,n}^2 + \epsilon} \cdot 
        \dfrac{\tilde{\boldsymbol{y}}_{t+1:t+H,n} - \beta_n}{\gamma_n}
    \right) 
    + \mu_{t,n}
    \]
\end{enumerate}

\subsection{ResMLP}

The ResMLP is a residual multilayer perceptron module used in both the time-domain and frequency-domain branches of FreDN, with separate parameters. Given an input $Z_{\text{in}} \in \mathbb{R}^{d \times l_{\text{in}}}$, the output is computed as:
\[
Z_{\text{out}} = \mathrm{Linear}_{\text{out}}\left( \mathrm{MLP}(Z_{\text{in}}) + \mathrm{Linear}_{\text{res}}(Z_{\text{in}}) \right)
\]
\[
\quad Z_{\text{out}} \in \mathbb{R}^{d \times l_{\text{out}}}.
\]
Here, the MLP path consists of $K$ stacked feedforward blocks with widths $\{l_1, \dots, l_K\}$, where each block applies a linear transformationan, optionally LayerNorm on even-indexed layers, GELU activation, and dropout. The residual path is a single linear layer mapping from $l_{\text{in}}$ directly to $l_K$. The sequence $\{l_1, \dots, l_K\}$ is generated via geometric interpolation from the hyperparameter \texttt{hidden\_size} to $l_{\text{out}}$, where $l_K$ is used as the unified hidden output dimension of the MLP. Finally, a linear projection maps the result to $l_{\text{out}}$. This is an empirical design to smoothly adjust the projection length, but does not play a central role in the model.

\subsection{MovDN and TopKDN}

We provide the details of the two decomposition baselines compared in our framework: MovDN and TopKDN.

\paragraph{MovDN.} MovDN follows the moving average decomposition strategy proposed in Autoformer~\cite{wu2021autoformer}, where the trend component is extracted via a symmetric padding and average pooling operation. Formally, given an input sequence $X \in \mathbb{R}^{C \times L \times d}$, the trend is obtained by applying a 1D average pooling over the temporal dimension with a fixed kernel size (we choose 25 following~\cite{wang2024timemixer}), and the seasonal component is computed as the residual between the original input and the smoothed trend. The implementation matches that of \texttt{series\_decomp} in the original Autoformer repository.

\paragraph{TopKDN.} TopKDN performs a frequency-domain decomposition by retaining only the dominant periodic patterns. Specifically, it first applies a real-valued FFT along the time axis to obtain the spectrum $\tilde{X} \in \mathbb{C}^{C \times L_{\text{freq}} \times d}$, where $L_{\text{freq}} = \lfloor L/2 \rfloor + 1$ is the lookback window length in the frequency domain. The amplitude spectrum is averaged across dimension $d$, and for each sample and variable, the top-$K$ frequencies with the highest magnitude are selected. An inverse FFT is then applied to obtain the seasonal part, while the residual is treated as the trend. To adapt to varying input lengths $L$, we set $K = \lfloor \log_2 L \rfloor$ as a heuristic design. The corresponding values are $\{6, 7, 8, 9, 9\}$ for $L\in \{96, 192, 336, 512, 720\}$.

\subsection{Complex-Linear Structure}
\label{appendix:CL}
\paragraph{Complex-Linear}
We provide the details of the Complex-Linear structure, in which all real-valued linear layers in ResMLP are replaced with complex-valued linear layers implemented via complex matrix multiplication. The residual formulation remains unchanged, but all linear mappings operate in the complex domain. Each complex linear layer applies a transformation to the input $\tilde{X} \in \mathbb{C}^{l_{\text{in}} \times d}$ using real-valued weight matrices. The weight and bias are parameterized as:
\begin{equation}
  \tilde{W} = \tilde{W}^{(r)} + j\tilde{W}^{(i)}, \quad \tilde{b} = \tilde{b}^{(r)} + j\tilde{b}^{(i)},
\end{equation}
where $\tilde{W}^{(r)}, \tilde{W}^{(i)} \in \mathbb{R}^{l_{\text{out}} \times l_{\text{in}}}$ and $\tilde{b}^{(r)}, \tilde{b}^{(i)} \in \mathbb{R}^{l_{\text{out}}}$. The output $\tilde{Y} = \tilde{W} \cdot \tilde{X} + \tilde{b}$ is computed by splitting the input into real and imaginary parts: $\tilde{X} = \tilde{X}^{(r)} + j \tilde{X}^{(i)}$, and applying the standard complex matrix rule:
\begin{equation}
\begin{aligned}
\mathrm{Re}(\tilde{Y}) &= \tilde{W}^{(r)} \cdot \tilde{X}^{(r)} - \tilde{W}^{(i)} \cdot \tilde{X}^{(i)} + \tilde{b}^{(r)}, \\
\mathrm{Im}(\tilde{Y}) &= \tilde{W}^{(i)} \cdot \tilde{X}^{(r)} + \tilde{W}^{(r)} \cdot \tilde{X}^{(i)} + \tilde{b}^{(i)}.
\end{aligned}
\end{equation}

In the Complex-Linear structure, multiple complex linear layers are stacked directly \emph{without} LayerNorm, GELU, or Dropout, as these operations are not natively defined for complex tensors in PyTorch.

\paragraph{Supplementary structures}
Trabelsi et al.~\cite{trabelsi2018deep} propose that activation and dropout can be applied separately to the real and imaginary parts. For LayerNorm, two design options are presented: (1) treating the complex tensor as a joint distribution and whitening it using the full covariance matrix between the real and imaginary parts; and (2) applying standard LayerNorm to the real and imaginary parts independently. 

To further evaluate the utility of our proposed ReIm Block, we introduce \textbf{Decorr-CL} and \textbf{Separate-CL} as enhanced variants of the Complex-Linear structure, which implement complex-valued LayerNorm, activation functions, and dropout mechanisms following the framework established by Trabelsi et al.~\cite{trabelsi2018deep}\footnote{\url{https://github.com/ChihebTrabelsi/deep_complex_networks}}. We present the comparative results in Appendix~\ref{appendix:Efficiencyfull}, ~\ref{appendix:suppCL}. 

\textbf{Decorr-CL} performs decorrelated normalization. Specifically, it computes the full second-order statistics of the real and imaginary components, normalizes using their joint covariance structure, and applies separate trainable weights and biases (i.e., $\gamma_{\mathrm{rr}}, \gamma_{\mathrm{ii}}, \gamma_{\mathrm{ri}}$, and corresponding $\beta$ terms). 

\textbf{Separate-CL} normalizes the real and imaginary parts independently using standard LayerNorm. 

In both cases, we follow the convention of Trabelsi et al.~\cite{trabelsi2018deep} where GELU activation and dropout are applied separately to the real and imaginary parts. The implementations used in our experiments mirrors our ResMLP exactly in topology, with Linear, LayerNorm, GELU, and dropout replaced by their complex counterparts as described above.

\section{Full Results}
\label{appendix:fullresults}
\subsection{Full Results of Forecasting}
The complete results of long-term forecasting on seven benchmarks are presented in Table~\ref{tab:mainfull}. Our proposed model, FreDN, ranks first in 30 out of 35 MSE entries and 32 out of 35 MAE entries. Notably, FreDN maintains top performance across a wide range of forecast horizons. 
\begin{table*}[htbp]
  \centering
  \scriptsize
  \setlength{\tabcolsep}{2pt}
  \caption{Full results of multivariate forecasting with horizon $H \in \{96, 192, 336, 720\}$ for all datasets.
We evaluate each model under multiple lookback window lengths $\{96, 192, 336, 512, 720\}$ and report the best performance. The best results are highlighted in \textbf{\textcolor{red}{bold red}}, and the second best are \underline{\textcolor{blue}{underlined blue}}.}
  \begin{adjustbox}{width=\textwidth}
    \begin{tabular}{lr|cc|cc|cc|cc|cc|cc|cc|cc}
    \toprule
    \multicolumn{2}{c|}{\multirow{1.5}[2]{*}{Models}} & \multicolumn{2}{c|}{FreDN} & \multicolumn{2}{c|}{SOFTS} & \multicolumn{2}{c|}{FreDF} & \multicolumn{2}{c|}{iTransformer} & \multicolumn{2}{c|}{TimeMixer} & \multicolumn{2}{c|}{FreTS} & \multicolumn{2}{c|}{PatchTST} & \multicolumn{2}{c}{DLinear} \\
    \multicolumn{2}{c|}{} & \multicolumn{2}{c|}{(Ours)} & \multicolumn{2}{c|}{(2024)} & \multicolumn{2}{c|}{(2025)} & \multicolumn{2}{c|}{(2024)} & \multicolumn{2}{c|}{(2023)} & \multicolumn{2}{c|}{(2023)} & \multicolumn{2}{c|}{(2023)} & \multicolumn{2}{c}{(2023)} \\
    \midrule
    \multicolumn{2}{c|}{Metrics} & MSE   & MAE   & MSE   & MAE   & MSE   & MAE   & MSE   & MAE   & MSE   & MAE   & MSE   & MAE   & MSE   & MAE   & MSE   & MAE \\
    \midrule
     \multicolumn{1}{c}{\multirow{5}[0]{*}{\rotatebox{90}{ETTh1}}} & 96    & \textcolor[rgb]{ 1,  0,  0}{\textbf{0.355 }} & \textcolor[rgb]{ 1,  0,  0}{\textbf{0.387 }} & 0.383  & 0.405  & 0.382  & 0.401  & 0.387  & 0.405  & 0.386  & 0.401  & 0.395  & 0.410  & 0.378  & 0.409  & \underline{\textcolor[rgb]{ 0,  0,  1}{0.367 }} & \underline{\textcolor[rgb]{ 0,  0,  1}{0.395 }} \\
          & 192   & \textcolor[rgb]{ 1,  0,  0}{\textbf{0.391 }} & \textcolor[rgb]{ 1,  0,  0}{\textbf{0.410 }} & 0.476  & 0.466  & 0.414  & 0.429  & 0.427  & 0.436  & 0.421  & 0.430  & 0.436  & 0.442  & 0.417  & 0.430  & \underline{\textcolor[rgb]{ 0,  0,  1}{0.400 }} & \underline{\textcolor[rgb]{ 0,  0,  1}{0.417 }} \\
          & 336   & \textcolor[rgb]{ 1,  0,  0}{\textbf{0.412 }} & \textcolor[rgb]{ 1,  0,  0}{\textbf{0.426 }} & 0.517  & 0.492  & 0.438  & 0.443  & 0.449  & 0.459  & 0.439  & 0.453  & 0.476  & 0.467  & 0.438  & 0.445  & \underline{\textcolor[rgb]{ 0,  0,  1}{0.430 }} & \underline{\textcolor[rgb]{ 0,  0,  1}{0.438 }} \\
          & 720   & \textcolor[rgb]{ 1,  0,  0}{\textbf{0.429 }} & \textcolor[rgb]{ 1,  0,  0}{\textbf{0.457 }} & 0.552  & 0.529  & \underline{\textcolor[rgb]{ 0,  0,  1}{0.452 }} & \underline{\textcolor[rgb]{ 0,  0,  1}{0.470 }} & 0.492  & 0.492  & 0.476  & 0.475  & 0.547  & 0.532  & 0.464  & 0.475  & 0.468  & 0.489  \\
          & Avg   & \textcolor[rgb]{ 1,  0,  0}{\textbf{0.397 }} & \textcolor[rgb]{ 1,  0,  0}{\textbf{0.420 }} & 0.482  & 0.473  & 0.421  & 0.436  & 0.438  & 0.448  & 0.430  & 0.440  & 0.463  & 0.463  & 0.424  & 0.440  & \underline{\textcolor[rgb]{ 0,  0,  1}{0.416 }} & \underline{\textcolor[rgb]{ 0,  0,  1}{0.435 }} \\
    \midrule
    \multicolumn{1}{c}{\multirow{5}[0]{*}{\rotatebox{90}{ETTh2}}} & 96    & \textcolor[rgb]{ 1,  0,  0}{\textbf{0.266 }} & \textcolor[rgb]{ 1,  0,  0}{\textbf{0.328 }} & 0.293  & 0.343  & 0.285  & 0.348  & 0.299  & 0.350  & \underline{\textcolor[rgb]{ 0,  0,  1}{0.280 }} & 0.343  & 0.312  & 0.373  & 0.282  & \underline{\textcolor[rgb]{ 0,  0,  1}{0.341 }} & 0.290  & 0.357  \\
          & 192   & \textcolor[rgb]{ 1,  0,  0}{\textbf{0.320 }} & \textcolor[rgb]{ 1,  0,  0}{\textbf{0.368 }} & 0.364  & 0.398  & 0.355  & 0.393  & 0.377  & 0.399  & 0.351  & 0.389  & 0.392  & 0.428  & \underline{\textcolor[rgb]{ 0,  0,  1}{0.350 }} & \underline{\textcolor[rgb]{ 0,  0,  1}{0.385 }} & 0.399  & 0.430  \\
          & 336   & \textcolor[rgb]{ 1,  0,  0}{\textbf{0.352 }} & \textcolor[rgb]{ 1,  0,  0}{\textbf{0.396 }} & 0.396  & 0.429  & 0.388  & 0.418  & 0.415  & 0.432  & 0.380  & 0.419  & 0.491  & 0.489  & \underline{\textcolor[rgb]{ 0,  0,  1}{0.373 }} & \underline{\textcolor[rgb]{ 0,  0,  1}{0.404 }} & 0.480  & 0.481  \\
          & 720   & \textcolor[rgb]{ 1,  0,  0}{\textbf{0.383 }} & \textcolor[rgb]{ 1,  0,  0}{\textbf{0.426 }} & 0.429  & 0.450  & 0.415  & 0.445  & 0.416  & 0.443  & 0.425  & 0.445  & 0.653  & 0.577  & \underline{\textcolor[rgb]{ 0,  0,  1}{0.398 }} & \underline{\textcolor[rgb]{ 0,  0,  1}{0.432 }} & 0.762  & 0.620  \\
          & Avg   & \textcolor[rgb]{ 1,  0,  0}{\textbf{0.330 }} & \textcolor[rgb]{ 1,  0,  0}{\textbf{0.380 }} & 0.371  & 0.405  & 0.361  & 0.401  & 0.377  & 0.406  & 0.359  & 0.399  & 0.462  & 0.467  & \underline{\textcolor[rgb]{ 0,  0,  1}{0.351 }} & \underline{\textcolor[rgb]{ 0,  0,  1}{0.391 }} & 0.483  & 0.472  \\
    \midrule
    \multicolumn{1}{c}{\multirow{5}[0]{*}{\rotatebox{90}{ETTm1}}} & 96    & \textcolor[rgb]{ 1,  0,  0}{\textbf{0.284 }} & \textcolor[rgb]{ 1,  0,  0}{\textbf{0.334 }} & 0.303  & 0.348  & 0.303  & 0.353  & 0.302  & 0.353  & 0.300  & 0.347  & 0.307  & 0.351  & 0.304  & 0.350  & \underline{\textcolor[rgb]{ 0,  0,  1}{0.300 }} & \underline{\textcolor[rgb]{ 0,  0,  1}{0.343 }} \\
          & 192   & \textcolor[rgb]{ 1,  0,  0}{\textbf{0.319 }} & \textcolor[rgb]{ 1,  0,  0}{\textbf{0.361 }} & 0.338  & 0.369  & 0.336  & 0.376  & 0.345  & 0.378  & 0.340  & 0.374  & 0.348  & 0.377  & 0.349  & 0.377  & \underline{\textcolor[rgb]{ 0,  0,  1}{0.335 }} & \underline{\textcolor[rgb]{ 0,  0,  1}{0.365 }} \\
          & 336   & \textcolor[rgb]{ 1,  0,  0}{\textbf{0.354 }} & \textcolor[rgb]{ 1,  0,  0}{\textbf{0.382 }} & \underline{\textcolor[rgb]{ 0,  0,  1}{0.367 }} & 0.390  & 0.371  & 0.396  & 0.379  & 0.403  & 0.374  & 0.397  & 0.383  & 0.399  & 0.378  & 0.392  & 0.370  & \underline{\textcolor[rgb]{ 0,  0,  1}{0.388 }} \\
          & 720   & \textcolor[rgb]{ 1,  0,  0}{\textbf{0.402 }} & \textcolor[rgb]{ 1,  0,  0}{\textbf{0.412 }} & 0.492  & 0.474  & 0.428  & 0.430  & 0.439  & 0.437  & 0.435  & 0.425  & 0.444  & 0.437  & 0.422  & 0.418  & \underline{\textcolor[rgb]{ 0,  0,  1}{0.415 }} & \underline{\textcolor[rgb]{ 0,  0,  1}{0.416 }} \\
          & Avg   & \textcolor[rgb]{ 1,  0,  0}{\textbf{0.340 }} & \textcolor[rgb]{ 1,  0,  0}{\textbf{0.372 }} & 0.375  & 0.395  & 0.360  & 0.389  & 0.366  & 0.393  & 0.362  & 0.386  & 0.370  & 0.391  & 0.363  & 0.384  & \underline{\textcolor[rgb]{ 0,  0,  1}{0.355 }} & \underline{\textcolor[rgb]{ 0,  0,  1}{0.378 }} \\
          \midrule
    \multicolumn{1}{c}{\multirow{5}[0]{*}{\rotatebox{90}{ETTm2}}} & 96    & \textcolor[rgb]{ 1,  0,  0}{\textbf{0.156 }} & \textcolor[rgb]{ 1,  0,  0}{\textbf{0.244 }} & \underline{\textcolor[rgb]{ 0,  0,  1}{0.164 }} & 0.254  & 0.175  & 0.258  & 0.175  & 0.266  & 0.171  & 0.257  & 0.172  & 0.261  & 0.165  & 0.256  & 0.166  & \underline{\textcolor[rgb]{ 0,  0,  1}{0.250 }} \\
          & 192   & \textcolor[rgb]{ 1,  0,  0}{\textbf{0.211 }} & \textcolor[rgb]{ 1,  0,  0}{\textbf{0.283 }} & 0.226  & 0.296  & 0.244  & 0.306  & 0.238  & 0.312  & 0.227  & 0.298  & 0.242  & 0.311  & 0.225  & 0.297  & \underline{\textcolor[rgb]{ 0,  0,  1}{0.221 }} & \underline{\textcolor[rgb]{ 0,  0,  1}{0.291 }} \\
          & 336   & \textcolor[rgb]{ 1,  0,  0}{\textbf{0.264 }} & \textcolor[rgb]{ 1,  0,  0}{\textbf{0.318 }} & 0.277  & 0.333  & 0.295  & 0.341  & 0.289  & 0.341  & 0.279  & 0.331  & 0.308  & 0.356  & 0.274  & 0.328  & \underline{\textcolor[rgb]{ 0,  0,  1}{0.273 }} & \underline{\textcolor[rgb]{ 0,  0,  1}{0.328 }} \\
          & 720   & \textcolor[rgb]{ 1,  0,  0}{\textbf{0.341 }} & \textcolor[rgb]{ 1,  0,  0}{\textbf{0.370 }} & 0.369  & 0.388  & 0.379  & 0.396  & 0.377  & 0.397  & 0.364  & 0.383  & 0.373  & 0.404  & \underline{\textcolor[rgb]{ 0,  0,  1}{0.354 }} & \underline{\textcolor[rgb]{ 0,  0,  1}{0.383 }} & 0.362  & 0.393  \\
          & Avg   & \textcolor[rgb]{ 1,  0,  0}{\textbf{0.243 }} & \textcolor[rgb]{ 1,  0,  0}{\textbf{0.304 }} & 0.259  & 0.318  & 0.273  & 0.325  & 0.270  & 0.329  & 0.260  & 0.317  & 0.274  & 0.333  & \underline{\textcolor[rgb]{ 0,  0,  1}{0.254 }} & \underline{\textcolor[rgb]{ 0,  0,  1}{0.316 }} & 0.255  & \underline{\textcolor[rgb]{ 0,  0,  1}{0.316 }} \\
          \midrule
    \multicolumn{1}{c}{\multirow{5}[0]{*}{\rotatebox{90}{Weather}}} & 96    & \textcolor[rgb]{ 1,  0,  0}{\textbf{0.139 }} & \textcolor[rgb]{ 1,  0,  0}{\textbf{0.186 }} & 0.156  & 0.206  & 0.162  & 0.213  & 0.163  & 0.213  & \underline{\textcolor[rgb]{ 0,  0,  1}{0.150 }} & \underline{\textcolor[rgb]{ 0,  0,  1}{0.204 }} & 0.151  & 0.210  & 0.159  & 0.211  & 0.168  & 0.228  \\
          & 192   & \textcolor[rgb]{ 1,  0,  0}{\textbf{0.183 }} & \textcolor[rgb]{ 1,  0,  0}{\textbf{0.230 }} & 0.204  & 0.250  & 0.207  & 0.254  & 0.203  & 0.249  & \underline{\textcolor[rgb]{ 0,  0,  1}{0.195 }} & \underline{\textcolor[rgb]{ 0,  0,  1}{0.244 }} & 0.195  & 0.253  & 0.203  & 0.251  & 0.211  & 0.266  \\
          & 336   & \textcolor[rgb]{ 1,  0,  0}{\textbf{0.232 }} & \textcolor[rgb]{ 1,  0,  0}{\textbf{0.270 }} & 0.251  & 0.287  & 0.259  & 0.293  & 0.255  & 0.290  & 0.245  & \underline{\textcolor[rgb]{ 0,  0,  1}{0.281 }} & \underline{\textcolor[rgb]{ 0,  0,  1}{0.242 }} & 0.293  & 0.258  & 0.293  & 0.259  & 0.310  \\
          & 720   & \textcolor[rgb]{ 1,  0,  0}{\textbf{0.305 }} & \textcolor[rgb]{ 1,  0,  0}{\textbf{0.324 }} & 0.322  & 0.337  & 0.338  & 0.345  & 0.326  & 0.338  & 0.323  & \underline{\textcolor[rgb]{ 0,  0,  1}{0.335 }} & \underline{\textcolor[rgb]{ 0,  0,  1}{0.309 }} & 0.340  & 0.320  & 0.339  & 0.314  & 0.351  \\
          & Avg   & \textcolor[rgb]{ 1,  0,  0}{\textbf{0.215 }} & \textcolor[rgb]{ 1,  0,  0}{\textbf{0.253 }} & 0.233  & 0.270  & 0.242  & 0.276  & 0.237  & 0.272  & 0.228  & \underline{\textcolor[rgb]{ 0,  0,  1}{0.266 }} & \underline{\textcolor[rgb]{ 0,  0,  1}{0.224 }} & 0.274  & 0.235  & 0.274  & 0.238  & 0.289  \\
    \midrule
    \multicolumn{1}{c}{\multirow{5}[0]{*}{\rotatebox{90}{Electricity}}} & 96    & \textcolor[rgb]{ 1,  0,  0}{\textbf{0.126 }} & \textcolor[rgb]{ 1,  0,  0}{\textbf{0.219 }} & \underline{\textcolor[rgb]{ 0,  0,  1}{0.130 }} & \underline{\textcolor[rgb]{ 0,  0,  1}{0.226 }} & \underline{\textcolor[rgb]{ 0,  0,  1}{0.130 }} & \underline{\textcolor[rgb]{ 0,  0,  1}{0.226 }} & 0.131  & 0.227  & 0.133  & \underline{\textcolor[rgb]{ 0,  0,  1}{0.226 }} & 0.135  & 0.275  & 0.133  & 0.270  & 0.137  & 0.286  \\
          & 192   & \textcolor[rgb]{ 1,  0,  0}{\textbf{0.144 }} & \textcolor[rgb]{ 1,  0,  0}{\textbf{0.236 }} & \underline{\textcolor[rgb]{ 0,  0,  1}{0.147 }} & \underline{\textcolor[rgb]{ 0,  0,  1}{0.241 }} & 0.149  & 0.245  & 0.153  & 0.249  & 0.154  & 0.245  & 0.151  & 0.288  & 0.149  & 0.278  & 0.151  & 0.289  \\
          & 336   & \textcolor[rgb]{ 1,  0,  0}{\textbf{0.159 }} & \textcolor[rgb]{ 1,  0,  0}{\textbf{0.252 }} & \underline{\textcolor[rgb]{ 0,  0,  1}{0.163 }} & \underline{\textcolor[rgb]{ 0,  0,  1}{0.257 }} & 0.166  & 0.262  & 0.169  & 0.265  & 0.169  & 0.264  & 0.168  & 0.304  & 0.168  & 0.283  & 0.166  & 0.296  \\
          & 720   & \textcolor[rgb]{ 1,  0,  0}{\textbf{0.191 }} & \textcolor[rgb]{ 1,  0,  0}{\textbf{0.283 }} & 0.199  & 0.290  & \underline{\textcolor[rgb]{ 0,  0,  1}{0.197 }} & 0.293  & \textcolor[rgb]{ 1,  0,  0}{\textbf{0.191 }} & \underline{\textcolor[rgb]{ 0,  0,  1}{0.285 }} & 0.206  & 0.296  & 0.208  & 0.334  & 0.207  & 0.303  & 0.201  & 0.318  \\
          & Avg   & \textcolor[rgb]{ 1,  0,  0}{\textbf{0.155 }} & \textcolor[rgb]{ 1,  0,  0}{\textbf{0.247 }} & \underline{\textcolor[rgb]{ 0,  0,  1}{0.160 }} & \underline{\textcolor[rgb]{ 0,  0,  1}{0.253 }} & 0.161  & 0.256  & 0.161  & 0.257  & 0.165  & 0.258  & 0.166  & 0.300  & 0.164  & 0.283  & 0.164  & 0.297  \\
    \midrule
    \multicolumn{1}{c}{\multirow{5}[0]{*}{\rotatebox{90}{Traffic}}} & 96    & 0.353  & \underline{\textcolor[rgb]{ 0,  0,  1}{0.247 }} & 0.375  & 0.269  & \textcolor[rgb]{ 1,  0,  0}{\textbf{0.348 }} & \textcolor[rgb]{ 1,  0,  0}{\textbf{0.243 }} & \underline{\textcolor[rgb]{ 0,  0,  1}{0.351 }} & 0.257  & 0.359  & 0.251  & 0.391  & 0.275  & 0.374  & 0.270  & 0.399  & 0.286  \\
          & 192   & 0.370  & \textcolor[rgb]{ 1,  0,  0}{\textbf{0.255 }} & 0.395  & 0.281  & \textcolor[rgb]{ 1,  0,  0}{\textbf{0.362 }} & \textcolor[rgb]{ 1,  0,  0}{\textbf{0.255 }} & \underline{\textcolor[rgb]{ 0,  0,  1}{0.365 }} & 0.265  & 0.371  & 0.261  & 0.412  & 0.288  & 0.389  & 0.278  & 0.408  & 0.289  \\
          & 336   & \underline{\textcolor[rgb]{ 0,  0,  1}{0.381 }} & \textcolor[rgb]{ 1,  0,  0}{\textbf{0.259 }} & 0.418  & 0.296  & \textcolor[rgb]{ 1,  0,  0}{\textbf{0.378 }} & \underline{\textcolor[rgb]{ 0,  0,  1}{0.263 }} & 0.384  & 0.273  & 0.388  & 0.265  & 0.435  & 0.304  & 0.404  & 0.283  & 0.422  & 0.296  \\
          & 720   & 0.420  & \underline{\textcolor[rgb]{ 0,  0,  1}{0.283 }} & 0.456  & 0.316  & \textcolor[rgb]{ 1,  0,  0}{\textbf{0.414 }} & \textcolor[rgb]{ 1,  0,  0}{\textbf{0.279 }} & \underline{\textcolor[rgb]{ 0,  0,  1}{0.415 }} & 0.287  & 0.427  & 0.291  & 0.488  & 0.334  & 0.442  & 0.303  & 0.462  & 0.318  \\
          & Avg   & 0.381  & \underline{\textcolor[rgb]{ 0,  0,  1}{0.261 }} & 0.411  & 0.290  & \textcolor[rgb]{ 1,  0,  0}{\textbf{0.376 }} & \textcolor[rgb]{ 1,  0,  0}{\textbf{0.260 }} & \underline{\textcolor[rgb]{ 0,  0,  1}{0.379 }} & 0.270  & 0.386  & 0.267  & 0.431  & 0.300  & 0.402  & 0.283  & 0.423  & 0.297  \\
    \midrule
    \multicolumn{2}{c|}{1st Count} & 30    & 32    & 0     & 0     & 5     & 4     & 1     & 0     & 0     & 0     & 0     & 0     & 0     & 0     & 0     & 0 \\
    \midrule
    \multicolumn{2}{c|}{2nd Count} & 1     & 3     & 6     & 4     & 3     & 3     & 4     & 1     & 3     & 6     & 3     & 0     & 6     & 7     & 10    & 13 \\
    \bottomrule
    \end{tabular}%
\end{adjustbox}
  \label{tab:mainfull}%
\end{table*}%

We perform \textbf{Wilcoxon signed-rank tests} on the forecasting errors across all 28 tasks (4 prediction lengths for  7 datasets) to assess whether the improvements of our model are statistically significant. As shown in Table~\ref{tab:wilcoxon}, FreDN significantly outperforms both FreDF and SOFTS ($p < 0.001$), confirming the robustness of its improvements. 
\begin{table*}[htbp]
\centering
\small
\vspace{-1em}
\caption{Wilcoxon signed-rank test results comparing FreDN with FreDF and SOFTS on 28 forecasting tasks. Statistically significant results ($p < 0.05$) are shown in bold.}
\label{tab:wilcoxon}
\renewcommand{\arraystretch}{1.2}
\begin{tabular}{c c c c c}
\toprule
\textbf{Comparison} & \textbf{Metric} & \textbf{Wilcoxon Statistic} & \textbf{p-value} & \textbf{Significant} \\
\midrule
FreDN vs FreDF & MSE & 0.0   & \textbf{3.7e-09} & \textbf{Yes} \\
FreDN vs SOFTS & MSE & 18.0  & \textbf{9.4e-07} & \textbf{Yes} \\
FreDN vs FreDF & MAE & 0.0   & \textbf{3.7e-09} & \textbf{Yes} \\
FreDN vs SOFTS & MAE & 3.5   & \textbf{4.1e-06} & \textbf{Yes} \\
\bottomrule
\end{tabular}
\end{table*}

\subsection{Comparison with TimeMixer++}
The comparison between FreDN and TimeMixer++ is provided in Table~\ref{tab:TM}. Although our best results with lookback window tuning outperform TimeMixer++, we do not include it in our main result comparison for fairness, as TimeMixer++ uses a fixed lookback window ($L=96$) and does not provide a publicly available official implementation. 
\begin{table*}[htbp]
  \vspace{-1em}
  \caption{Comparison with TimeMixer++ (\textbf{bold} denotes better results). Our FreDN consistently outperforms TimeMixer++ across all datasets.}
  \resizebox{\textwidth}{!}{
    \centering
    \small
    \begin{tabular}{c|c|>{\centering\arraybackslash}p{4em}|>{\centering\arraybackslash}p{4em}|>{\centering\arraybackslash}p{4em}|>{\centering\arraybackslash}p{4em}|>{\centering\arraybackslash}p{4em}|>{\centering\arraybackslash}p{4em}|>{\centering\arraybackslash}p{4em}}
    \toprule
    \multicolumn{1}{c|}{Models} & Metrics & ETTh1 & ETTh2 & ETTm1 & ETTm2 & Weather & Electricity & Traffic \\
    \midrule
    FreDN & MSE   & \textbf{0.397 } & \textbf{0.330 } & \textbf{0.340 } & \textbf{0.243 } & \textbf{0.215 } & \textbf{0.155 } & \textbf{0.381 } \\
    (Ours) & MAE   & \textbf{0.420 } & 0.380  & \textbf{0.372 } & \textbf{0.304 } & \textbf{0.253 } & \textbf{0.247 } & \textbf{0.261 } \\
    \midrule
    TimeMixer++ & MSE   & 0.419  & 0.339  & 0.369  & 0.269  & 0.226  & 0.165  & 0.416  \\
    (2025) & MAE   & 0.432  & 0.380  & 0.378  & 0.320  & 0.262  & 0.253  & 0.264  \\
    \bottomrule
    \end{tabular}%
  }
  \label{tab:TM}%
  \vspace{-2em}
\end{table*}%

\subsection{Full Results of Different Lookback Window Length}
\begin{figure}[t]
  \centering
  \begin{subfigure}{0.48\linewidth}
    \includegraphics[width=\linewidth]{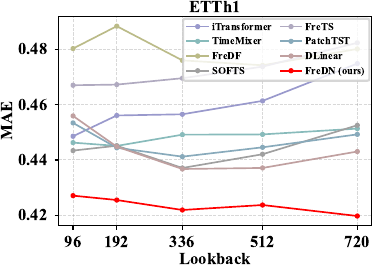}
    \caption{ETTh1}
  \end{subfigure}
  \hfill
  \begin{subfigure}{0.48\linewidth}
    \includegraphics[width=\linewidth]{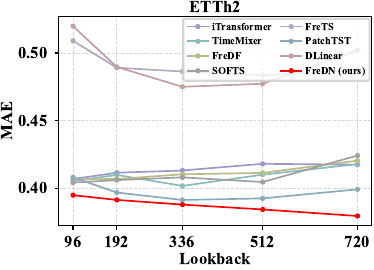}
    \caption{ETTh2}
  \end{subfigure}
  \caption{Effect of lookback window length on the average MAE across all prediction horizons for ETTh1 and ETTh2.}
  \label{fig:lookbackfull0}
\end{figure}

\begin{figure}[t]
  \centering
  \begin{subfigure}{0.48\linewidth}
    \includegraphics[width=\linewidth]{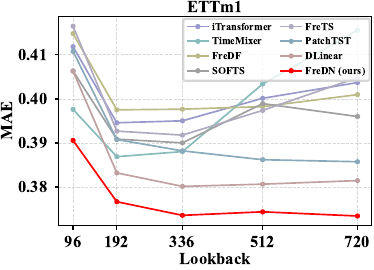}
    \caption{ETTm1}
  \end{subfigure}
  \hfill
  \begin{subfigure}{0.48\linewidth}
    \includegraphics[width=\linewidth]{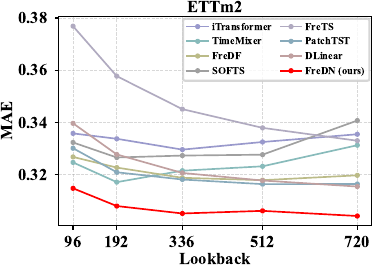}
    \caption{ETTm2}
  \end{subfigure}
  \caption{Effect of lookback window length on the average MAE across all prediction horizons for ETTm1 and ETTm2.}
  \label{fig:lookbackfull1}
\end{figure}

\begin{figure}[t]
  \centering
  \begin{subfigure}{0.32\linewidth}
    \includegraphics[width=\linewidth]{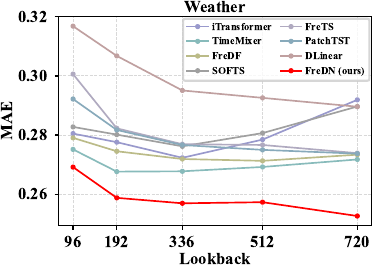}
    \caption{Weather}
  \end{subfigure}
  \hfill
  \begin{subfigure}{0.32\linewidth}
    \includegraphics[width=\linewidth]{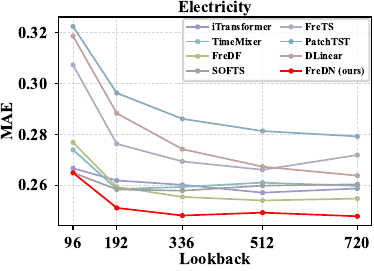}
    \caption{Electricity}
  \end{subfigure}
  \hfill
  \begin{subfigure}{0.32\linewidth}
    \includegraphics[width=\linewidth]{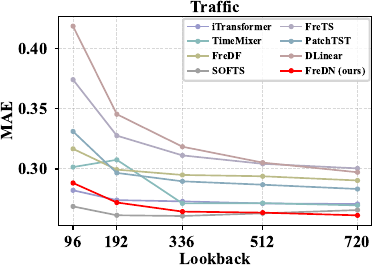}
    \caption{Traffic}
  \end{subfigure}
  \caption{Effect of lookback window length on the average MAE across all prediction horizons for Weather, Electricity, and Traffic.}
  \label{fig:lookbackfull2}
\end{figure}

Figures~\ref{fig:lookbackfull0}, \ref{fig:lookbackfull1}, \ref{fig:lookbackfull2} illustrate the impact of lookback window length on prediction accuracy across all seven datasets. For each lookback length, we report the average MAE over four forecasting horizons ($\tau \in \{96, 192, 336, 720\}$) as the comparison metric. As shown, most baseline models tend to suffer from performance degradation when the lookback window increases, except on the Electricity and Traffic datasets. In contrast, FreDN consistently improves as the input length grows. On the high-dimensional datasets Electricity and Traffic, where longer inputs tend to be beneficial for most models, FreDN still achieves the steepest error reduction. These results highlight the effectiveness of FreDN in leveraging the information gain from extended historical contexts.

\subsection{Full Results of Efficiency Analysis}\label{appendix:Efficiencyfull}
\begin{figure}[t]
  \centering
  \includegraphics[width=0.8\linewidth]{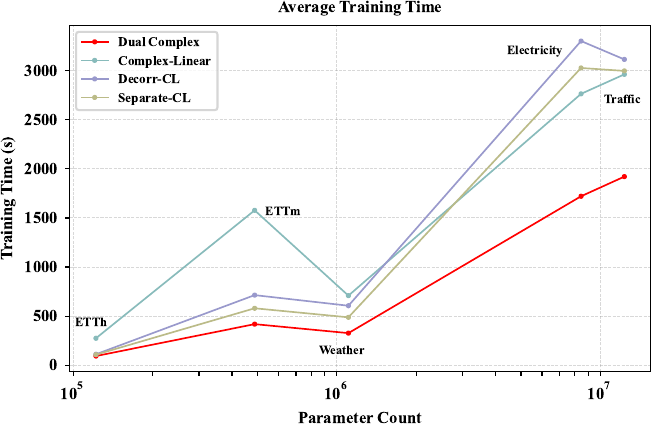}
  \caption{Average running time of different structures. ETTh and ETTm represent the averages over ETTh1/ETTh2 and ETTm1/ETTm2, respectively.}
  \label{fig:Efficiencyfull}
\end{figure}

We provide additional runtime comparisons for the two alternative complex-valued structures introduced in Appendix~\ref{appendix:CL}: Decorr-CL and Separate-CL. Both variants are built upon the base Complex-Linear structure, augmented with different LayerNorm mechanisms, complex-valued activation functions and dropout layers. As a result, they retain nearly identical parameter counts to Complex-Linear structure, and exhibit similar computational efficiency in practice. The average runtime results are summarized in Figure~\ref{fig:Efficiencyfull}.

\subsection{Full Results of Hyperparameter Sensitivity Analysis}
\begin{figure}[t]
  \centering
  \begin{subfigure}{0.48\linewidth}
    \includegraphics[width=\linewidth]{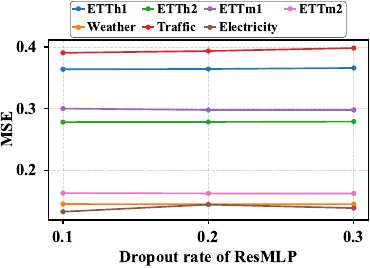}
    \caption{Dropout rate in ResMLPs}
  \end{subfigure}
  \hfill
  \begin{subfigure}{0.48\linewidth}
    \includegraphics[width=\linewidth]{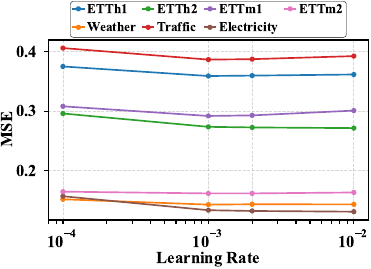}
    \caption{Learning rate}
  \end{subfigure}
  \caption{Influence of hyperparameters including the dropout rate and learning rate.}
  \label{fig:sensitivityfull0}
\end{figure}

\begin{figure}[t]
  \centering
  \begin{subfigure}{0.48\linewidth}
    \includegraphics[width=\linewidth]{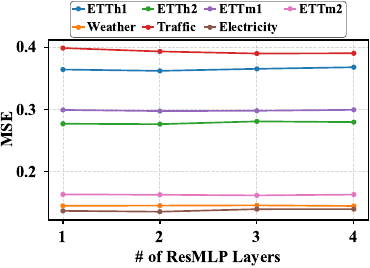}
    \caption{Number of hidden layers}
  \end{subfigure}
  \hfill
  \begin{subfigure}{0.48\linewidth}
    \includegraphics[width=\linewidth]{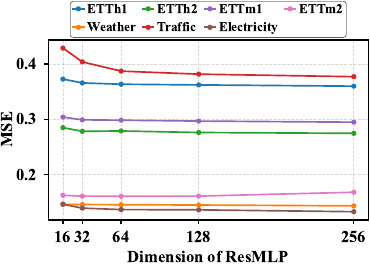}
    \caption{Hidden dimension}
  \end{subfigure}
  \caption{Influence of hyperparameters including the number of ResMLP hidden layers and the hidden dimension.}
  \label{fig:sensitivityfull1}
\end{figure}

In this section, we present the full results of the hyperparameter sensitivity analysis for FreDN, including four key factors: the hidden dimension of ResMLP, the number of layers, the learning rate, and the dropout rate applied within ResMLPs. As shown in Figure~\ref{fig:sensitivityfull0}, \ref{fig:sensitivityfull1}, datasets with higher feature dimensionality, such as Traffic and Electricity, benefit from wider and deeper ResMLPs, while lower-dimensional datasets like ETTs and Weather prefer more compact configurations, likely due to the strong feature compression provided by frequency-domain representations. In addition, we observe that the performance is relatively stable across a wide range of learning rates, with $10^{-3}$ being consistently effective. The effect of dropout is minimal. These findings further demonstrate the robustness and adaptability of FreDN across diverse temporal settings.

\subsection{Supplementary Results of Complex-Linear Ablation}
In Table~\ref{tab:complexfull}, we report the full comparison results for the two enhanced Complex-Linear structures introduced in Appendix~\ref{appendix:CL}: Decorr-CL and Separate-CL. These two structures extend the base Complex-Linear structure by incorporating complex-valued LayerNorm, activation functions, and dropout, following the formulation proposed by Trabelsi et al.~\cite{trabelsi2018deep}. Specifically, Decorr-CL computes the full second-order statistics of the real and imaginary components, normalizes using their joint covariance structure, and applies separate trainable affine parameters. This makes Decorr-CL heavier in terms of parameter count compared to the base Complex-Linear. 

Despite these enhancements, neither Decorr-CL nor Separate-CL exhibits significant performance improvement over Complex-Linear. In contrast, our proposed ReIm Block maintains competitive accuracy while reducing the parameter count by over 50\%, highlighting its efficiency.

\label{appendix:suppCL}
\begin{table*}[htbp]
  \centering
  \caption{Performance comparison between Dual-Complex, Complex-Linear, Decorr-CL, and Separate-CL. The reported results are averaged over four prediction lengths $\tau \in \{96, 192, 336, 720\}$.}
  \resizebox{\textwidth}{!}{
    \small
    \begin{tabular}{c|c|>{\centering\arraybackslash}p{4em}|>{\centering\arraybackslash}p{4em}|>{\centering\arraybackslash}p{4em}|>{\centering\arraybackslash}p{4em}|>{\centering\arraybackslash}p{4em}|>{\centering\arraybackslash}p{4em}|>{\centering\arraybackslash}p{4em}}
    \toprule
    Structures & Metrics & ETTh1 & ETTh2 & ETTm1 & ETTm2 & Weather & Electricity & Traffic \\
    \midrule
    \multirow{2}[2]{*}{Dual Complex} & MSE   & \textbf{0.397 } & \textbf{0.330 } & \textbf{0.340 } & \textbf{0.243 } & \textbf{0.215 } & \textbf{0.155 } & \textbf{0.381 } \\
\cmidrule{2-9}          & MAE   & \textbf{0.420 } & \textbf{0.380 } & \textbf{0.372 } & \textbf{0.304 } & \textbf{0.253 } & \textbf{0.247 } & \textbf{0.261 } \\
    \midrule
    \multirow{2}[2]{*}{Complex-Linear} & MSE   & 0.403  & 0.344  & 0.345  & 0.248  & 0.216  & 0.157  & 0.394  \\
\cmidrule{2-9}          & MAE   & 0.421  & 0.384  & \textbf{0.372 } & 0.305  & \textbf{0.253 } & 0.250  & 0.270  \\
    \midrule
    \multirow{2}[2]{*}{Decorr-CL} & MSE   & 0.411  & 0.342  & 0.348  & 0.249  & 0.242  & 0.158  & 0.384  \\
\cmidrule{2-9}          & MAE   & 0.431  & 0.386  & 0.376  & 0.308  & 0.275  & 0.251  & 0.263  \\
    \midrule
    \multirow{2}[2]{*}{Separate-CL} & MSE   & 0.412  & 0.341  & 0.347  & 0.249  & 0.223  & 0.156  & 0.384  \\
\cmidrule{2-9}          & MAE   & 0.430  & 0.385  & 0.376  & 0.308  & 0.261  & 0.248  & 0.264  \\
    \bottomrule
    \end{tabular}%
  }
  \label{tab:complexfull}%
\end{table*}%

\section{Proof}

We provide a complete justification of the theoretical claims. For each result, we explicitly state the assumptions, and apply standard tools such as Wirtinger calculus, matrix derivatives, and properties of linear transformations. 
\paragraph{Notation.}
Define the residual $\epsilon = \hat{Y} - Y$. Let $F$ be the DFT matrix and $F^H$ its Hermitian transpose. The frequency-domain residual is $\tilde{\epsilon} = F\cdot \epsilon \in \mathbb{C}^{C \times \tau_{\text{freq}}}$. Let $\tilde{X}_{\text{season}}, \tilde{Y}_{\text{season}} \in \mathbb{C}^{C \times \tau_{\text{freq}}}$ denote the input and output of the frequency-domain subnetwork. The output is further expressed as $\tilde{Y}_{\text{season}} = \tilde{Y}_{\text{season}}^{(r)} + j \tilde{Y}_{\text{season}}^{(i)}$. Let the loss function is $\mathcal{L}_F = \frac{1}{\tau_{\text{freq}}} \|\tilde{\epsilon} \|_1$.

\subsection{Proof of Theorem: Spectral of Sobolev-Smooth Trends}

\textbf{Spectral of Sobolev-Smooth Trends.}
Let \( f \in W^{m,2}([0,1]) \) be a Sobolev-smooth function with square-integrable $m$-th derivative for some \( m \geq 1 \). Let \( \hat{f}(k) \) be the \( k \)-th Fourier coefficient of the periodic extension of \( f \). Then there exists a constant \( C > 0 \) such that for all \( k \in \mathbb{Z} \setminus \{0\} \),
\[
0 \leq |\hat{f}(k)| \leq \frac{C}{|k|^m}.
\]

\textbf{Proof.}
Let \( f \in W^{m,2}([0,1]) \), which implies that the $m$-th weak derivative \( f^{(m)} \) exists and lies in \( L^2([0,1]) \). To analyze the decay of Fourier coefficients \( \hat{f}(k) \), we first consider the $k$-th coefficient of the periodic extension of \( f \), given by
\[
\hat{f}(k) = \int_0^1 f(x) e^{-2\pi i k x} \, dx.
\]

Since \( f^{(m)} \in L^2 \), and smooth periodic functions are dense in \( W^{m,2} \), we can integrate by parts $m$ times. Each integration by parts introduces a factor \( (2\pi i k)^{-1} \), and the boundary terms vanish due to periodicity. After $m$ such steps, we obtain:
\[
\hat{f}(k) = \frac{1}{(2\pi i k)^m} \int_0^1 f^{(m)}(x) e^{-2\pi i k x} \, dx.
\]

Taking absolute value and applying the Cauchy–Schwarz inequality:
\[
|\hat{f}(k)| \leq \frac{1}{(2\pi |k|)^m} \left\| f^{(m)} \right\|_{L^2([0,1])}.
\]

Define \( C := \left\| f^{(m)} \right\|_{L^2([0,1])} / (2\pi)^m \), which is finite since \( f^{(m)} \in L^2 \). Thus, for all \( k \neq 0 \),
\[
0 \leq |\hat{f}(k)| \leq \frac{C}{|k|^m},
\]
as claimed.
\qed

\subsection{Proof of Theorem: ReIm Block}
\label{sec:proof_reim}

\textbf{Theorem (ReIm Block).}  
Let \( \tilde{X} = [\tilde{X}_1, \dots, \tilde{X}_d]^\top \in \mathbb{C}^d \) be a complex-valued input vector.
\begin{itemize}
    \item \textbf{(Complex Linear Projection)} Let \( W \in \mathbb{C}^d \). Then the standard complex projection \( \tilde{Y}_{\text{complex}} = W^\top \tilde{X} \) can represent any complex number.
    \item \textbf{(ReIm Block)} Let \( W_r \in \mathbb{R}^d \). The projection \( \tilde{Y}_{\text{real}} = W_r^\top \tilde{X} \) can represent any complex number in \( \mathbb{C} \) if and only if there exist at least two entries \( \tilde{X}_i, \tilde{X}_j \) such that
    \[
    \arg(\tilde{X}_i) - \arg(\tilde{X}_j) \notin \pi \mathbb{Z}.
    \]
\end{itemize}

\textbf{Proof.}  
For the complex linear projection, since \( W \in \mathbb{C}^d \), we may choose arbitrary complex weights to match any desired output. Therefore, for any \( z \in \mathbb{C} \), there exists \( W \) such that \( W^\top \tilde{X} = z \), provided \( \tilde{X} \neq 0 \). Hence, full expressiveness is guaranteed.

For the ReIm Block, we consider the real-valued projection:
\[
\tilde{Y}_{\text{real}} = \sum_{i=1}^d w_i \tilde{X}_i, \quad w_i \in \mathbb{R}.
\]
This is a real linear combination of complex vectors \( \tilde{X}_i \in \mathbb{C} \). Therefore, the image of this projection lies in the real span \( \mathrm{span}_{\mathbb{R}} \{\tilde{X}_i\}_{i=1}^d \subseteq \mathbb{C} \), where we treat \(\mathbb{C}\) as a real vector space of dimension 2.

As shown in Hoffman and Kunze~\cite{hoffman1971linear}, two vectors \( \tilde{X}_i, \tilde{X}_j \in \mathbb{C} \) are linearly independent over \( \mathbb{R} \) if and only if their ratio is not real, i.e.,
\[
\frac{\tilde{X}_i}{\tilde{X}_j} \notin \mathbb{R} \quad \Leftrightarrow \quad \arg(\tilde{X}_i) - \arg(\tilde{X}_j) \notin \pi \mathbb{Z}.
\]
When such a pair exists, the set \( \{\tilde{X}_i\} \) spans \(\mathbb{C}\) as a real vector space, and thus any \( z \in \mathbb{C} \) can be represented as a real linear combination of the \(\tilde{X}_i\)'s.

Hence, the ReIm Block is fully expressive over \(\mathbb{C}\) if and only if the input contains at least two real-linearly independent complex numbers, which is equivalent to the phase difference condition.

\qed

\begin{figure}[t]
  \centering
  \includegraphics[width=0.8\linewidth]{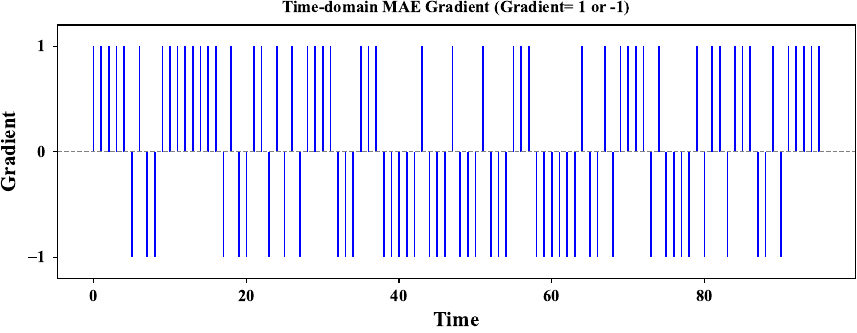}
  \caption{Illustration of time-domain MAE gradient. Each time step receives a unit gradient determined solely by the local residual sign.}
  \label{fig:timeMAE}
\end{figure}

\begin{figure}[t]
  \centering
  \includegraphics[width=0.8\linewidth]{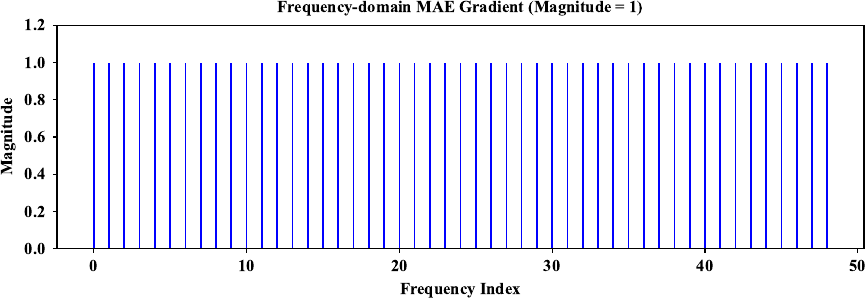}
  \caption{Illustration of ${\tilde{\epsilon}}/{|\tilde{\epsilon}|}$, which consists of unit-magnitude complex values in the frequency domain, each carrying phase information.}
  \label{fig:FreqMAE}
\end{figure}

\begin{figure}[t]
  \centering
  \includegraphics[width=0.8\linewidth]{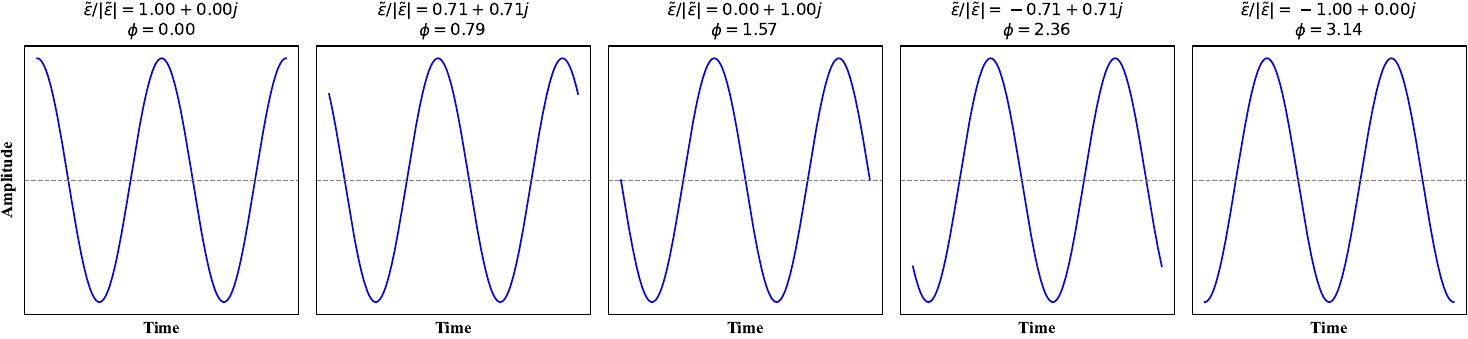}
  \caption{Effect of phase in frequency-domain gradient. Each subfigure shows the time-domain waveform corresponding to a single frequency with unit-magnitude and different phases.}
  \label{fig:phase}
\end{figure}

\begin{figure}[t]
  \centering
  \includegraphics[width=0.8\linewidth]{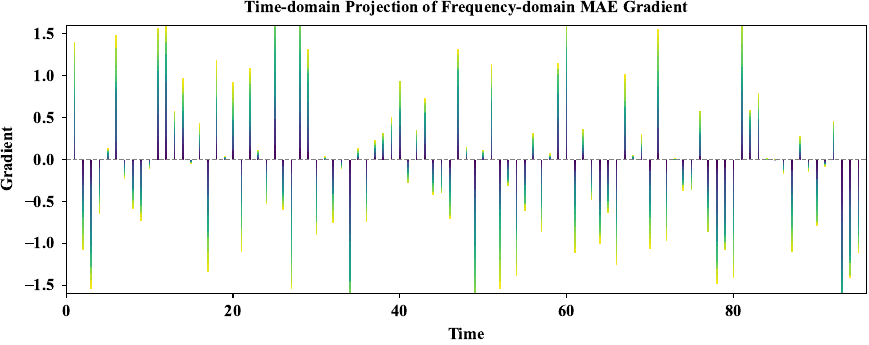}
  \caption{Time-domain projection of frequency-domain MAE gradient. Each vertical line is composed of contributions from multiple frequencies, revealing its global structure. Colors correspond to the proportional contributions of distinct frequency components at each temporal position.}
  \label{fig:FreqMAETime}
\end{figure}

\subsection{Gradient Derivation for Different Loss Functions}
We now provide detailed derivations for the gradients of four loss functions, focusing on their behavior with respect to the predicted output \( \hat{Y} \).

\paragraph{Time-domain MSE:}
\begin{align}
\mathcal{L}_{\text{time-MSE}} &= \frac{1}{\tau} \| \hat{Y} - Y\|_2^2 = \frac{1}{\tau} \| \epsilon \|_2^2, \\
\frac{\partial \mathcal{L}_{\text{time-MSE}}}{\partial \hat{Y}} &= \frac{2}{\tau} (\hat{Y} - Y) = \frac{2}{\tau} \epsilon.
\end{align}

\paragraph{Frequency-domain MSE:}
\begin{align}
\mathcal{L}_{\text{freq-MSE}} &= \frac{1}{\tau_{\text{freq}}} \| F(\hat{Y} - Y) \|_2^2 = \frac{1}{\tau_{\text{freq}}} \| \tilde{\epsilon} \|_2^2, \\
\frac{\partial \mathcal{L}_{\text{freq-MSE}}}{\partial \hat{Y}} &= \frac{2}{\tau_{\text{freq}}} F^H F (\hat{Y} - Y) = \frac{2}{\tau_{\text{freq}}} \epsilon.
\end{align}
Here, we use the unitarity of DFT: \( F^H F = I \).

\paragraph{Time-domain MAE:}
\begin{align}
\mathcal{L}_{\text{time-MAE}} &= \frac{1}{\tau} \|\hat{Y} - Y\|_1 = \frac{1}{\tau} \| \epsilon \|_1, \\
\frac{\partial \mathcal{L}_{\text{time-MAE}}}{\partial \hat{Y}} &= \frac{1}{\tau} \cdot \text{sign}(\epsilon),
\end{align}
where \( \text{sign}(x) \) is the element-wise sign function.

\paragraph{Frequency-domain MAE:}
\begin{align}
\mathcal{L}_{\text{freq-MAE}} &= \frac{1}{\tau_{\text{freq}}} \|F(\hat{Y} - Y)\|_1 = \frac{1}{\tau_{\text{freq}}} \| \tilde{\epsilon} \|_1 , \\
\frac{\partial \mathcal{L}_{\text{freq-MAE}}}{\partial \hat{Y}} &= \frac{\partial \mathcal{L}_{\text{freq-MAE}}}{\partial \tilde{\epsilon}} \cdot \frac{\partial \tilde{\epsilon}}{\partial \hat{Y}} = \frac{1}{\tau_{\text{freq}}} F^H \left( \frac{\tilde{\epsilon}}{|\tilde{\epsilon}|} \right).
\end{align}
Note that the division $\frac{\tilde{\epsilon}}{|\tilde{\epsilon}|}$ is performed element-wise across all frequency components. Therefore, the inverse DFT matrix $F^H$ cannot be moved inside the division operator. 
\qed
\paragraph{Interpretation of Gradients}
To illustrate the structural difference between the gradients of time-domain and frequency-domain MAE losses, we provide visual examples in Figures~\ref{fig:timeMAE}--\ref{fig:FreqMAETime}. As shown in Figure~\ref{fig:timeMAE}, the time-domain MAE produces local, isolated gradients: each time step receives a unit gradient determined solely by the residual sign. In contrast, Figure~\ref{fig:FreqMAE} shows that the key part of frequency-domain MAE gradients ${\tilde{\epsilon}}/{|\tilde{\epsilon}|}$ are complex-valued with unit magnitude, encoding global periodic information via phase. Figure~\ref{fig:phase} further demonstrates how phase affects the temporal structure of single-frequency components. The final synthesized gradient in Figure~\ref{fig:FreqMAETime} shows that each time-step update is composed of multiple frequency contributions, resulting in globally structured gradients. For simplicity, we omit the constant scalar factor in all derivations. This analysis supports that frequency-domain MAE enables global coordination during backpropagation.

\section{Error Bar}
We report the averaged results along with standard deviations of FreDN performance under five runs with different random seeds. The results on the four ETT datasets are summarized in Table~\ref{tab:errorbar0}, while those on the Weather, Electricity, and Traffic datasets are shown in Table~\ref{tab:errorbar1}. The consistently low standard deviations across all datasets and horizons demonstrate the stable and reliable performance of FreDN. 

\begin{table*}[htbp]
  \centering
  \caption{Error bars of FreDN for ETT datasets. Results are averaged over 5 experiments with different random seeds. }
  \resizebox{\textwidth}{!}{
    \begin{tabular}{c|c|c|c|c|c|c|c|c}
    \toprule
    Datasets & \multicolumn{2}{c|}{ETTh1} & \multicolumn{2}{c|}{ETTh2} & \multicolumn{2}{c|}{ETTm1} & \multicolumn{2}{c}{ETTm2} \\
    \midrule
    Horizon & MSE   & MAE   & MSE   & MAE   & MSE   & MAE   & MSE   & MAE \\
    \midrule
    96    & 0.357 $\pm$ 0.002 & 0.389 $\pm$ 0.001 & 0.268 $\pm$ 0.002 & 0.330 $\pm$ 0.001 & 0.286 $\pm$ 0.002 & 0.337 $\pm$ 0.002 & 0.161 $\pm$ 0.003 & 0.246 $\pm$ 0.002 \\
    192   & 0.392 $\pm$ 0.001 & 0.411 $\pm$ 0.000 & 0.327 $\pm$ 0.004 & 0.371 $\pm$ 0.002 & 0.326 $\pm$ 0.004 & 0.363 $\pm$ 0.002 & 0.215 $\pm$ 0.003 & 0.285 $\pm$ 0.002 \\
    336   & 0.418 $\pm$ 0.004 & 0.430 $\pm$ 0.001 & 0.354 $\pm$ 0.003 & 0.400 $\pm$ 0.003 & 0.355 $\pm$ 0.002 & 0.382 $\pm$ 0.001 & 0.267 $\pm$ 0.003 & 0.320 $\pm$ 0.002 \\
    720   & 0.434 $\pm$ 0.005 & 0.457 $\pm$ 0.004 & 0.387 $\pm$ 0.004 & 0.428 $\pm$ 0.003 & 0.408 $\pm$ 0.004 & 0.411 $\pm$ 0.001 & 0.344 $\pm$ 0.002 & 0.371 $\pm$ 0.001 \\
    \bottomrule
    \end{tabular}%
  }
  \label{tab:errorbar0}%
\end{table*}%

\begin{table*}[htbp]
  \centering
  \small
  \caption{Error bars of FreDN for Weather, Electricity, and Traffic datasets. Results are averaged over 5 experiments with different random seeds. }
  \resizebox{\textwidth}{!}{
    \begin{tabular}{c|c|c|c|c|c|c}
    \toprule
    Datasets & \multicolumn{2}{c|}{Weather} & \multicolumn{2}{c|}{Electricity} & \multicolumn{2}{c}{Traffic} \\
    \midrule
    Horizon & MSE   & MAE   & MSE   & MAE   & MSE   & MAE \\
    \midrule
    96    & 0.142 $\pm$ 0.002 & 0.188 $\pm$ 0.001 & 0.127 $\pm$ 0.000 & 0.220 $\pm$ 0.001 & 0.355 $\pm$ 0.001 & 0.248 $\pm$ 0.001 \\
    192   & 0.185 $\pm$ 0.001 & 0.231 $\pm$ 0.001 & 0.145 $\pm$ 0.001 & 0.237 $\pm$ 0.001 & 0.370 $\pm$ 0.000 & 0.255 $\pm$ 0.000 \\
    336   & 0.234 $\pm$ 0.002 & 0.272 $\pm$ 0.002 & 0.160 $\pm$ 0.001 & 0.254 $\pm$ 0.002 & 0.382 $\pm$ 0.001 & 0.261 $\pm$ 0.002 \\
    720   & 0.306 $\pm$ 0.004 & 0.326 $\pm$ 0.003 & 0.192 $\pm$ 0.001 & 0.283 $\pm$ 0.001 & 0.421 $\pm$ 0.001 & 0.283 $\pm$ 0.000 \\
    \bottomrule
    \end{tabular}%
  }
  \label{tab:errorbar1}%
\end{table*}%

\begin{figure}[t]
  \centering
  \includegraphics[width=\linewidth]{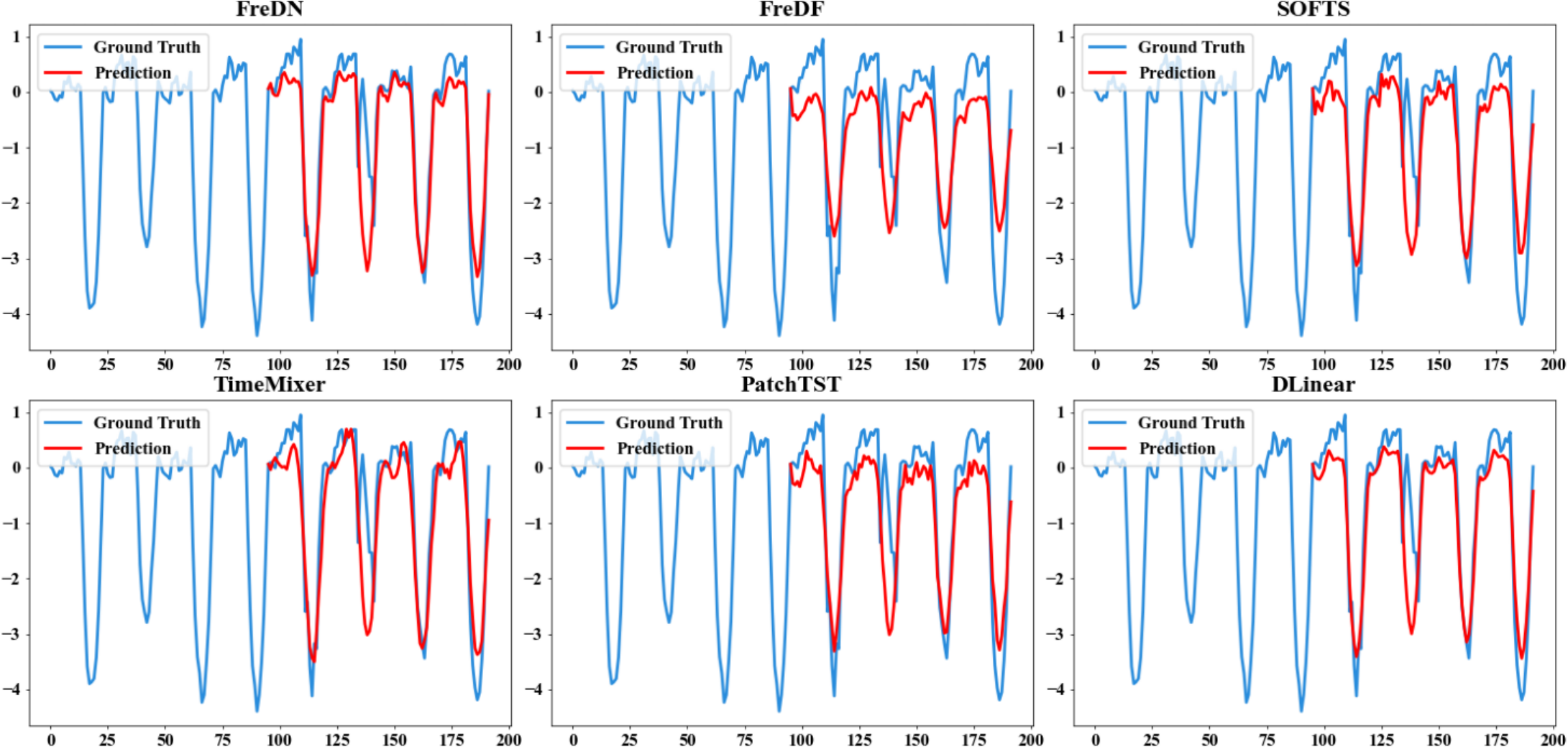}
  \caption{Forecasting visualization on the \textbf{ETTh1} dataset (input length $L=720$, prediction length $\tau=96$). To highlight the prediction window, we only plot the last 96 time steps of the lookback. Ground truth is shown in \textcolor{blue}{blue}, and predictions are shown in \textcolor{red}{red}.}
  \label{fig:ETTh1}
\end{figure}
\begin{figure}[t]
  \centering
  \includegraphics[width=\linewidth]{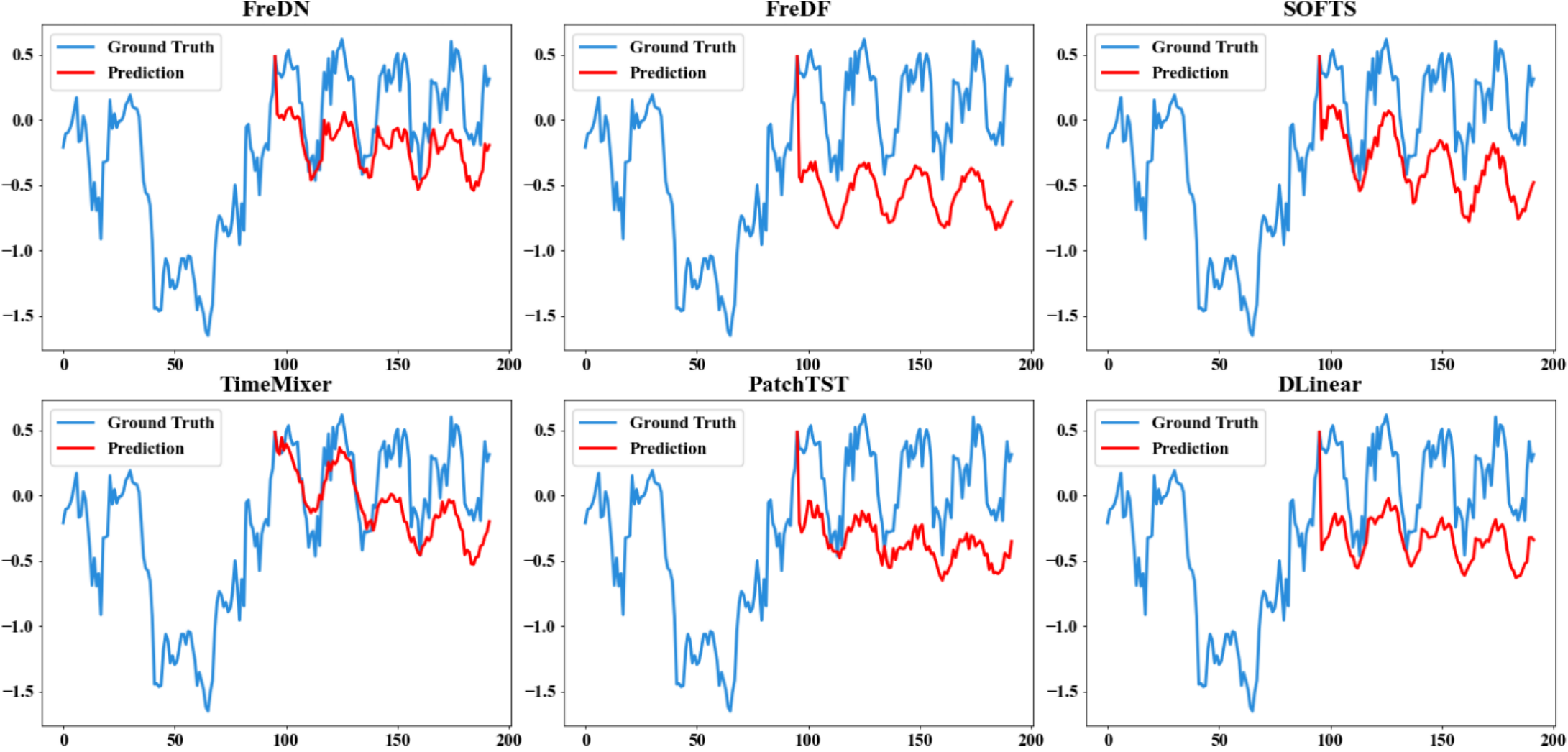}
  \caption{Forecasting visualization on the \textbf{ETTh2} dataset (input length $L=720$, prediction length $\tau=96$). To highlight the prediction window, we only plot the last 96 time steps of the lookback. Ground truth is shown in \textcolor{blue}{blue}, and predictions are shown in \textcolor{red}{red}.}
  \label{fig:ETTh2}
\end{figure}
\begin{figure}[t]
  \centering
  \includegraphics[width=\linewidth]{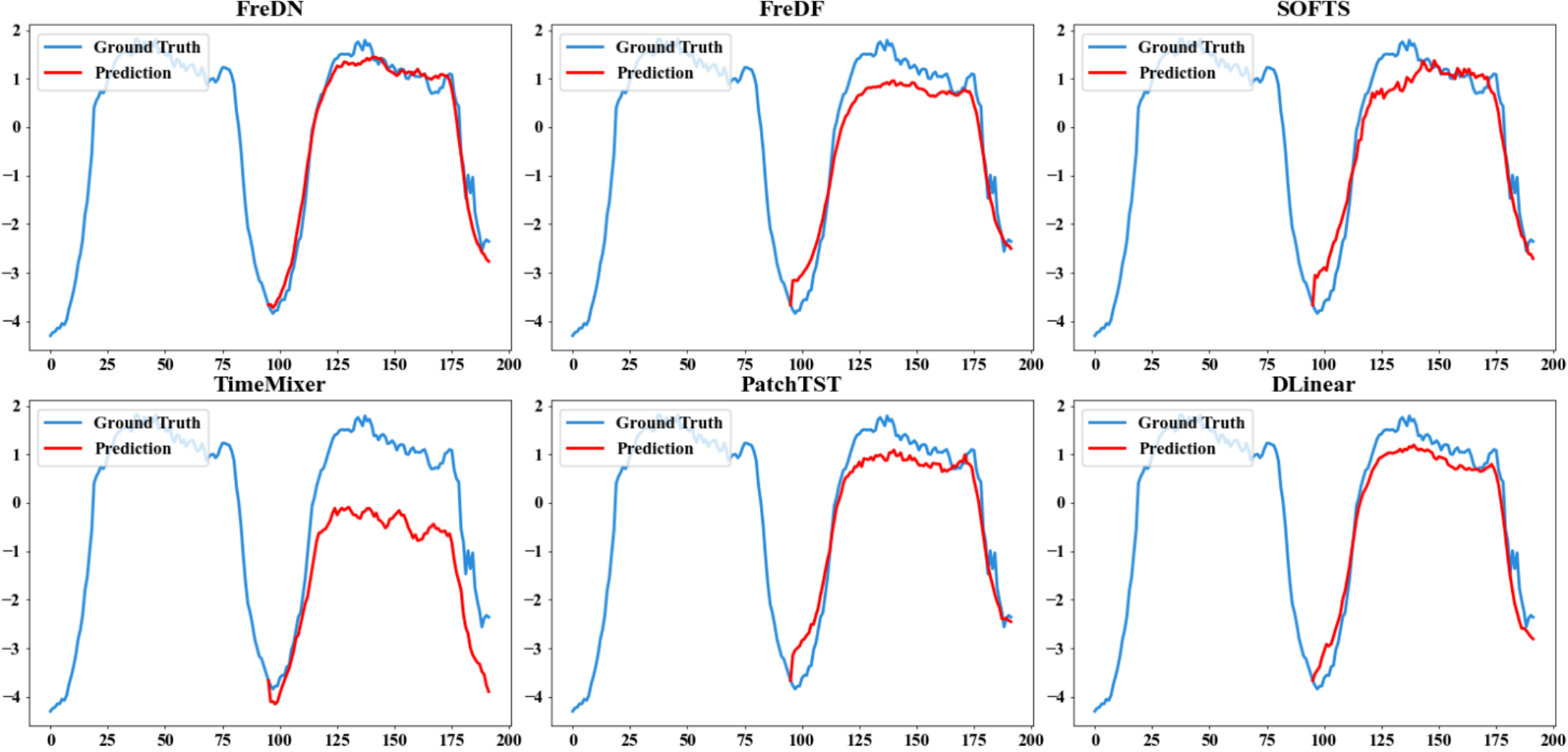}
  \caption{Forecasting visualization on the \textbf{ETTm1} dataset (input length $L=720$, prediction length $\tau=96$). To highlight the prediction window, we only plot the last 96 time steps of the lookback. Ground truth is shown in \textcolor{blue}{blue}, and predictions are shown in \textcolor{red}{red}.}
  \label{fig:ETTm1}
\end{figure}
\begin{figure}[t]
  \centering
  \includegraphics[width=\linewidth]{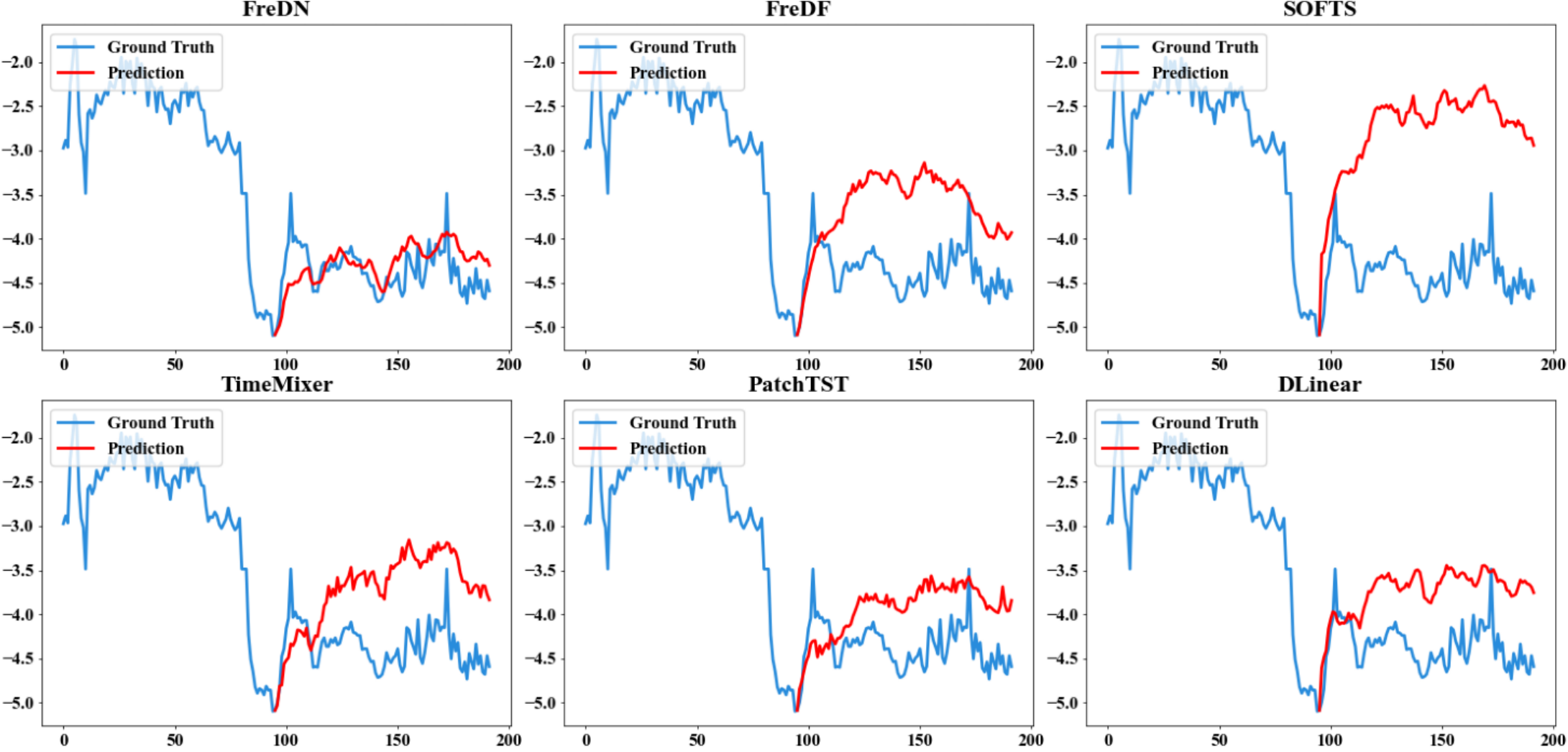}
  \caption{Forecasting visualization on the \textbf{ETTm2} dataset (input length $L=720$, prediction length $\tau=96$). To highlight the prediction window, we only plot the last 96 time steps of the lookback. Ground truth is shown in \textcolor{blue}{blue}, and predictions are shown in \textcolor{red}{red}.}
  \label{fig:ETTm2}
\end{figure}
\begin{figure}[t]
  \centering
  \includegraphics[width=\linewidth]{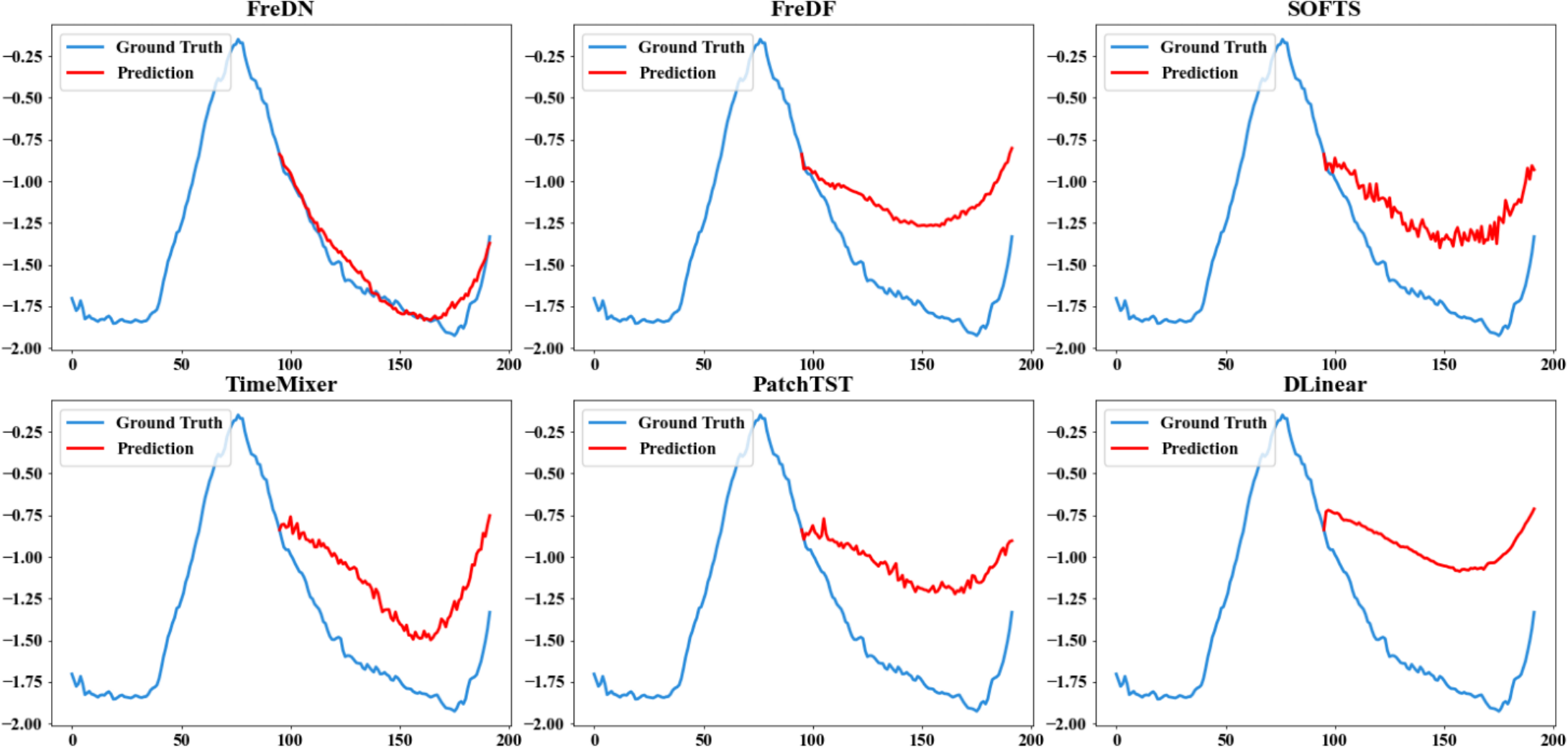}
  \caption{Forecasting visualization on the \textbf{weather} dataset (input length $L=720$, prediction length $\tau=96$). To highlight the prediction window, we only plot the last 96 time steps of the lookback. Ground truth is shown in \textcolor{blue}{blue}, and predictions are shown in \textcolor{red}{red}.}
  \label{fig:weather}
\end{figure}
\begin{figure}[t]
  \centering
  \includegraphics[width=\linewidth]{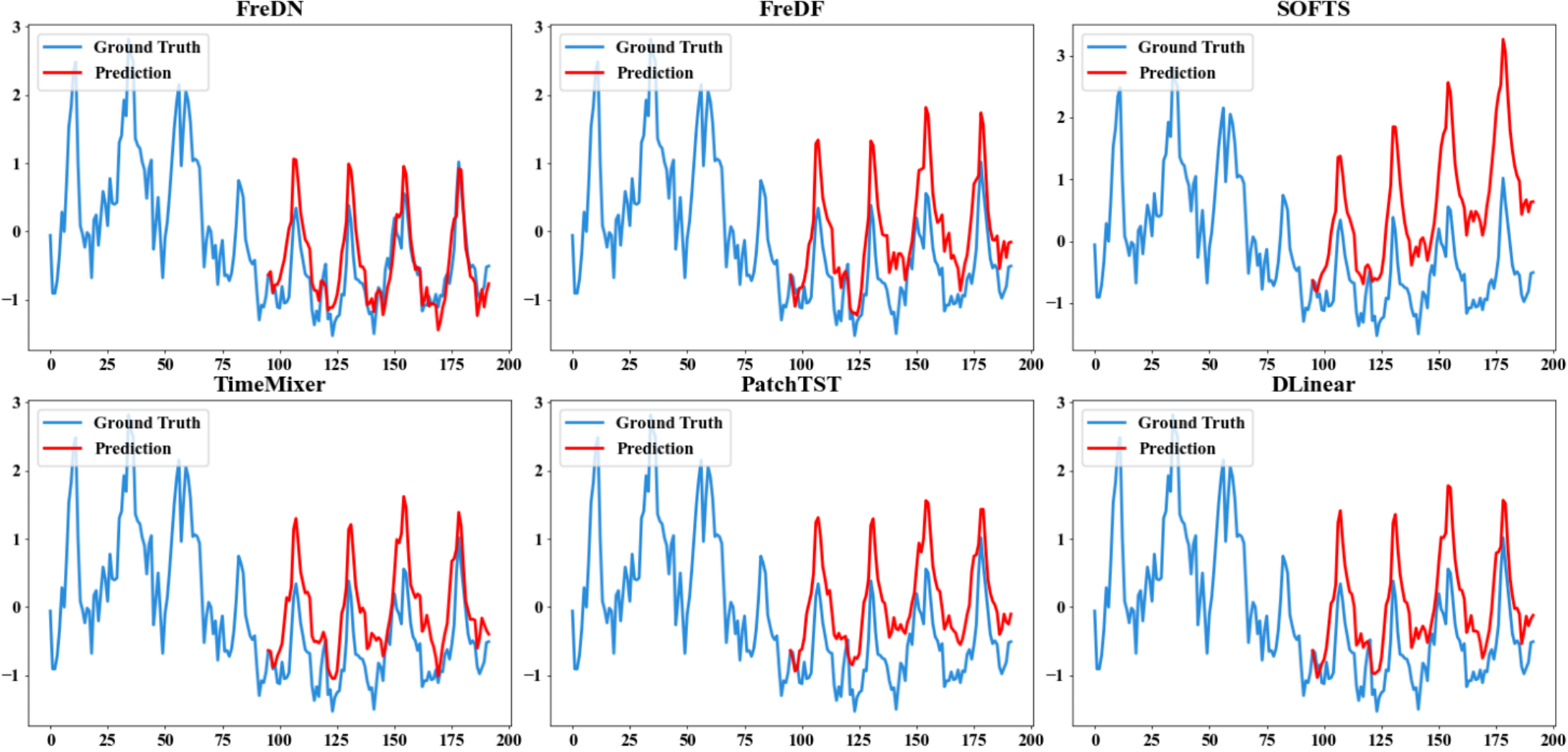}
  \caption{Forecasting visualization on the \textbf{Electricity} dataset (input length $L=720$, prediction length $\tau=96$). To highlight the prediction window, we only plot the last 96 time steps of the lookback. Ground truth is shown in \textcolor{blue}{blue}, and predictions are shown in \textcolor{red}{red}.}
  \label{fig:electricity}
\end{figure}
\begin{figure}[t]
  \centering
  \includegraphics[width=\linewidth]{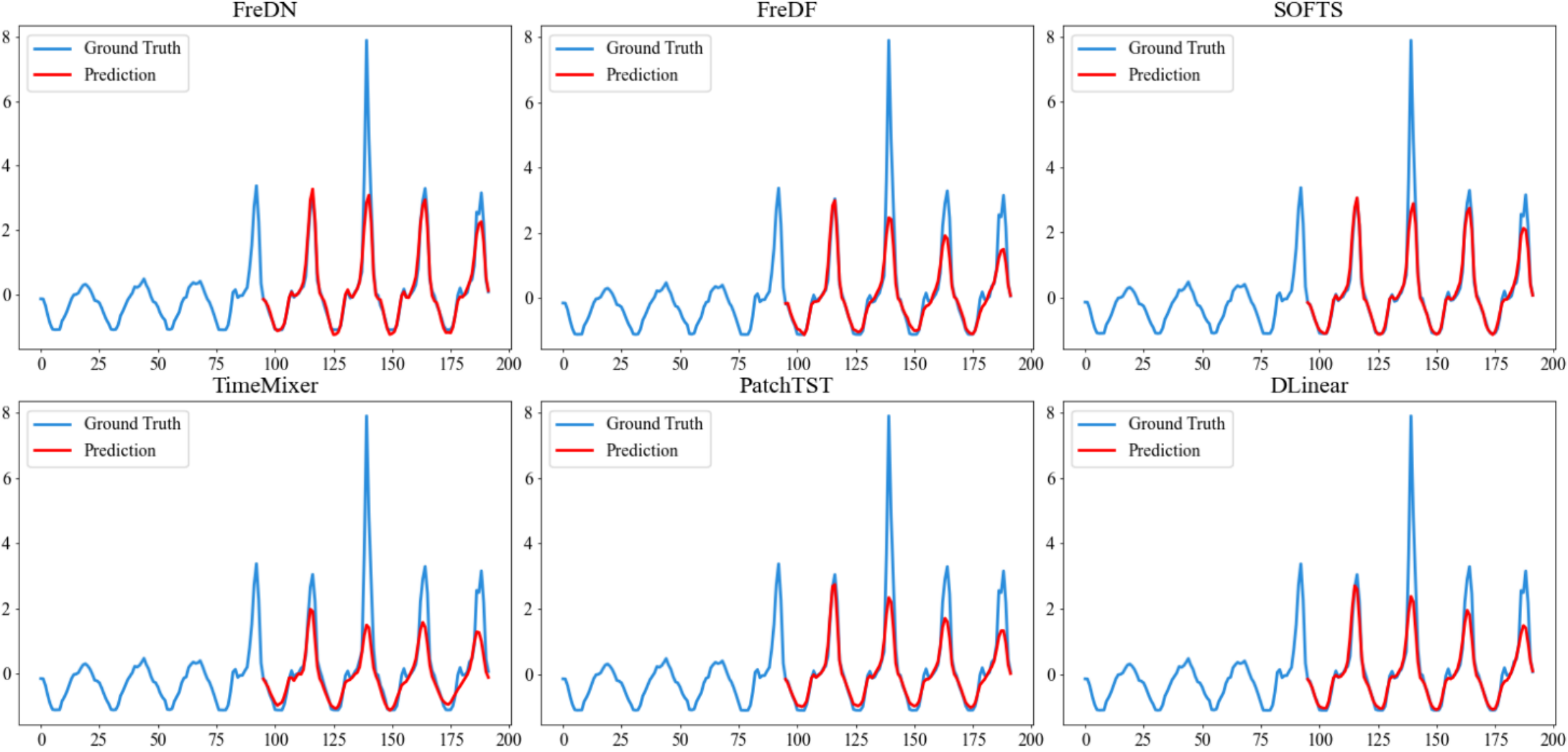}
  \caption{Forecasting visualization on the \textbf{Traffic} dataset (input length $L=720$, prediction length $\tau=96$). To highlight the prediction window, we only plot the last 96 time steps of the lookback. Ground truth is shown in \textcolor{blue}{blue}, and predictions are shown in \textcolor{red}{red}.}
  \label{fig:traffic}
\end{figure}

\section{Visualization}
W present supplementary prediction visualizations on all seven datasets in Figures~\ref{fig:ETTh1}, \ref{fig:ETTh2}, \ref{fig:ETTm1}, \ref{fig:ETTm2}, \ref{fig:weather}, \ref{fig:electricity}, \ref{fig:traffic}, covering the following representative baselines: FreDF~\cite{wang2025fredf}, SOFTS~\cite{han2024softs}, TimeMixer~\cite{wang2024timemixer}, PatchTST~\cite{nie2023patchtst}, and DLinear~\cite{zeng2023transformers}, alongside our proposed FreDN. To highlight the strength in leveraging long-range temporal information, we set the input sequence length to $L=720$. For layout clarity, we fix the prediction horizon to $\tau=96$ and visualize only the last $\tau$ time steps of the input. Among all methods, FreDN delivers the most coherent and accurate forecasts.
\section{Limitation and Future Work}

FreDN introduces a novel framework for long-term time series forecasting by decomposing multivariate signals in the frequency domain and modeling trend and seasonal components separately through  residual MLPs with ReIm Block. This design enables effective extraction of global periodic patterns while maintaining architectural simplicity. Nonetheless, several limitations and open challenges remain.

First, while our learnable Frequency Disentangler allows the model to isolate trend and seasonal components adaptively, it does not incorporate any explicit prior on periodicity or smoothness. Future work could explore combining signal processing priors or dynamic mode selection mechanisms to improve interpretability and generalization in more irregular series.

Second, our use of parameter-sharing ReIm Block provides a lightweight alternative to full complex-valued networks. However, more expressive complex modeling techniques, such as adaptive phase control, complex gating, or structured disentanglement, may further enhance the learning capacity and merit investigation.

Third, although FreDN demonstrates consistent improvements across diverse benchmarks, the benefit of frequency-domain modeling may vary depending on dataset characteristics such as input length, dimensionality, and periodic strength. Incorporating instance-adaptive strategies that selectively control the frequency emphasis or hybridize time-frequency learning may improve flexibility across tasks.

In summary, while FreDN offers an efficient and principled spectral modeling strategy for time series forecasting, further theoretical refinement and architectural extensions could improve its robustness, interpretability, and applicability to a wider range of real-world settings.

\section{Societal Impacts}

This work proposes a spectral decomposition framework for time series forecasting, which leverages Fourier transforms and ReIm Block to improve long-range prediction accuracy. Such models are applicable to a wide range of real-world domains, including energy planning, transportation systems, financial trend analysis, and environmental monitoring. The emphasis on interpretability and modular structure may also facilitate adoption in safety-critical scenarios.

While our model improves forecasting performance by capturing global periodic patterns, its use of Fourier-based decomposition introduces implicit assumptions about the input signal. Specifically, the discrete Fourier transform treats the input as a finite segment of a periodic signal and assumes that the underlying structure is sufficiently smooth to be represented by harmonic components. These assumptions may not hold in domains with highly irregular, bursty, or nonstationary data, which could affect the reliability of the learned representations.

Therefore, in practical deployment, we recommend combining FreDN with safeguards such as residual diagnostics, uncertainty quantification, and periodic model recalibration. Attention should also be paid to identifying spurious or misleading spectral structures, especially when working with complex multivariate systems. Domain knowledge and continuous evaluation remain critical to ensure responsible and trustworthy use of spectral forecasting methods.

\end{document}